\pdfoutput=1
\documentclass[11pt]{article}

\usepackage[final]{acl}

\usepackage{times}
\usepackage{latexsym}

\usepackage[T1]{fontenc}
\usepackage[utf8]{inputenc}

\usepackage{microtype}

\usepackage{inconsolata}

\usepackage{graphicx}

\usepackage{amsmath}
\usepackage{amsfonts} % add for mathbb
\usepackage{tcolorbox}
\usepackage{algorithm}
\usepackage[noend]{algpseudocode}
\usepackage{tabularray} % For the tblr environment
\usepackage{tcolorbox}  % For colored boxes and styling
\usepackage{xcolor}     % For text coloring
\usepackage{graphicx} % for including images
\usepackage{caption}  % for caption outside float environments
\usepackage{multirow} 
\usepackage{amsmath}
\usepackage{array}
\usepackage{makecell}
\usepackage{booktabs}
\usepackage{enumitem}
\usepackage{wrapfig}
\usepackage{subcaption}

\title{Understanding Impact of Human Feedback via Influence Functions}

\author{Taywon Min$^{1}$,
  Haeone Lee$^{1}$,
  Yongchan Kwon$^{2}$, Kimin Lee$^{1}$ \\ [1ex]
  ${}^{1}$KAIST; ${}^{2}$Columbia University \\ [1ex]
  \texttt{kiminlee@kaist.ac.kr}
}

\newtheorem{theorem}{Theorem}[section]
\newtheorem{remark}[theorem]{Remark}

\newcommand{\tw}[1]{\textcolor{red}{#1}}

\newcommand{\incinfluence}{\textit{negatively-contributing}}
\newcommand{\decinfluence}{\textit{positively-contributing}}
\begin{document}
\maketitle
% In Reinforcement Learning from Human Feedback (RLHF), it is crucial to learn suitable reward models from human feedback to align large language models (LLMs) with human intentions. However, human feedback can often be noisy, inconsistent, or biased, especially when evaluating complex responses. Such feedback can lead to misaligned reward signals, potentially causing unintended side effects during the RLHF process. To address these challenges, we explore the use of influence functions to measure the impact of human feedback on the performance of reward models. We propose a compute-efficient approximation method that enables the application of influence functions to LLM-based reward models and large-scale preference datasets. In our experiments, we demonstrate two key applications of influence functions: (1) detecting common forms of labeler bias in human feedback datasets and (2) guiding labelers to refine their strategies to align more closely with expert feedback. By quantifying the impact of human feedback, we believe that influence functions can enhance feedback interpretability and contribute to scalable oversight in RLHF, helping labelers provide more accurate and consistent feedback.
\begin{abstract}
In Reinforcement Learning from Human Feedback (RLHF), it is crucial to learn suitable reward models from human feedback to align large language models (LLMs) with human intentions. However, human feedback can often be noisy, inconsistent, or biased, especially when evaluating complex responses. Such feedback can lead to misaligned reward signals, potentially causing unintended side effects during the RLHF process. To address these challenges, we explore the use of influence functions to measure the impact of human feedback on the performance of reward models. We propose a compute-efficient approximation method that enables the application of influence functions to LLM-based reward models and large-scale preference datasets. 
Our experiments showcase two key applications of influence functions: (1) detecting common labeler biases in human feedback datasets and (2) guiding labelers in refining their strategies to better align with expert feedback.
% \tw{In our experiments, we demonstrate two key applications of influence functions. First, we test our method's ability to detect common forms of labeler bias on human feedback datasets with small portions of labels modified to introduce bias, then extend our analysis to real datasets using a human survey. \yc{We can add a strong statement like (Our results demonstrate that XXX or XX \% of data points in XXXX dataset is potentially mislabeled.)} Second, we guide labelers to refine their strategies to better align with expert feedback.} 
By quantifying the impact of human feedback, we believe that influence functions can enhance feedback interpretability and contribute to scalable oversight in RLHF, helping labelers provide more accurate and consistent feedback. Source code is available at \href{https://github.com/mintaywon/IF\_RLHF}{\texttt{https://github.com/mintaywon/IF\_RLHF}}.
\end{abstract}

% First, we identify common forms of labeler bias in human feedback datasets, then extend our analysis to real datasets using human surveys.
% First, we test our method's ability to detect common forms of labeler bias on human feedback datasets where a small fraction of preference labels have been flipped to introduce bias, then extend our analysis to real datasets using human surveys.

\section{Introduction}

As large language models (LLMs) demonstrate remarkable capabilities across various domains, ensuring their behaviors align with human intentions becomes increasingly important.
To this end, reinforcement learning from human feedback (RLHF) has emerged as a powerful solution for fine-tuning LLMs~\citep{ziegler2019fine,stiennon2020learning,instructgpt}.
In RLHF, human feedback is collected to train reward models that capture important human values, such as helpfulness and harmlessness~\citep{bai2022hh, ji2024beavertails}.
LLMs are then fine-tuned to produce outputs that closely align with these reward models.

However, human feedback can often be noisy, inconsistent, or biased, especially when evaluating complex responses~\citep{rlhfchallenges}. 
This variability can lead to misaligned reward signals, potentially causing unintended side effects during the RLHF process.
For example, feedback that favors supportive and enthusiastic responses might inadvertently lead the reward model to prioritize overly agreeable responses, which could result in sycophantic behavior~\citep{sharma2023towardssyco, perez2022discovering}.
This issue highlights the need for robust methods that precisely evaluate the impact of feedback on reward models, enabling humans to detect biased feedback and refine their feedback strategies more effectively.

% In this work, we investigate how to measure the impact of human feedback on the performance of reward models using the influence function~\citep{hampel1974influence, koh2017understanding}.
In this work, we assess the impact of human feedback on reward models by utilizing influence functions~\citep{hampel1974influence, koh2017understanding}.
%\tw{In this work, we apply influence functions~\citep{hampel1974influence, koh2017understanding} to assess the effect of human feedback on reward models, allowing for more precise evaluation by quantifying feedback’s impact on model outcomes. However, a}
%However, a major bottleneck in applying influence functions to reward models, especially those with large parameters such as LLMs and extensive preference datasets, is the high computational cost.
However, a significant challenge arises when applying influence functions to reward models, especially large-parameter models like LLMs and those involving extensive preference datasets, due to the high computational costs involved.
To address this, we introduce a compute-efficient method that utilizes vector compression techniques~\citep{li2023oporp} alongside the influence estimation method~\citep{kwon2023datainf}, achieving a 2.5-fold speed acceleration compared to previous methods in computing influence functions.
This approach significantly reduces the computational costs required to compute influence functions, facilitating more practical applications in large-scale settings.

\begin{figure*}[t!]
    \centering
    \includegraphics[width=0.88\textwidth]{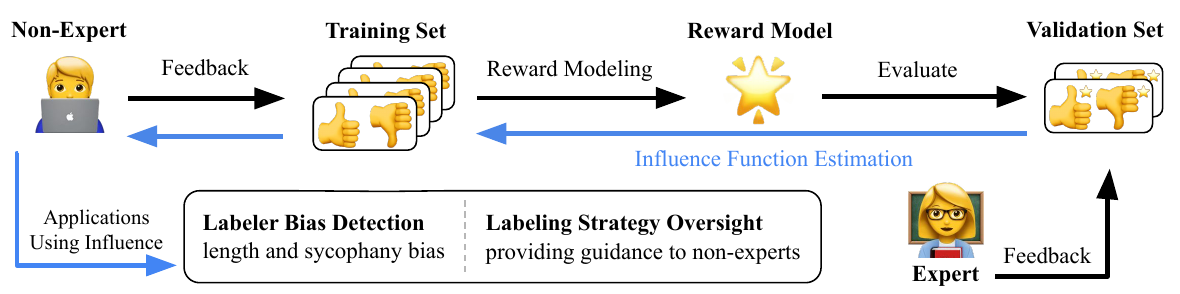}
    \captionof{figure}{An overview of our work, which applies influence functions to reward modeling. We apply influence functions to critical tasks such as labeler bias detection and labeling strategy oversight, enhancing the interpretability of human feedback in RLHF.}
    \label{fig:intro_labler}
\end{figure*}

We demonstrate two applications of influence functions (see \autoref{fig:intro_labler} for an overview): (1) detecting labeler bias in training datasets, and (2) improving suboptimal labeling strategies. 
In our first experiment, we explore two prevalent biases in the RLHF paradigm: length and sycophancy bias, where labelers may naively prefer longer~\citep{saito2023verbosity} and more sycophantic responses~\citep{sharma2023towardssyco}, regardless of response quality.
% originally method
To evaluate our approach, we introduce labeler bias by flipping a small portion of the Anthropic-HH dataset~\cite{bai2022hh} and assess whether they can be detected using influence functions.
Our approach significantly outperforms several baselines, including GPT-4o~\citep{gpt4o} and various outlier detection methods~\citep{lee2018maha, sun2022knn}, by effectively identifying biased samples.
We further extend our analysis to original Anthropic-HH datasets (i.e., without flipping) using human surveys, confirming its robustness. 
Specifically, when analyzing 100 samples flagged by our approach, human annotators found 47 mislabeled cases, indicating the prevalence of labeler bias in real-world datasets.
These results demonstrate the potential of our method to enhance publicly available preference datasets by accurately detecting and mitigating labeler biases.

% \tw{We further extend our analysis to real datasets using human surveys, confirming that influence functions remain effective in detecting labeler biases in real datasets. Specifically, when analyzing 100 samples identified by our approach, human annotators found 47 samples from the length dataset and 55 samples from the sycophancy dataset to be mislabeled, indicating that labeler bias is prevalent in real datasets. This further highlights our method's usefulness for enhancing the quality of publicly available preference datasets through accurate bias detection.}

Additionally, we showcase the utility of influence functions in refining feedback strategies to better align with expert evaluations using a proof-of-concept experiment.
% In a proof-of-concept experiment using the Helpsteer 2 dataset~\citep{wang2024helpsteer2}, we simulate a scenario with an expert labeler (Alice) using optimal weights and a non-expert labeler (Bob) using suboptimal weights. 
% By analyzing the influence scores of the validation set labeled by Alice, we effectively identify the training samples where Bob’s labeling deviates from Alice’s optimal strategy. 
% This insight allows Bob to adjust his weights to better match the expert’s, enhancing the accuracy of his evaluations.
Utilizing the Helpsteer 2 dataset~\citep{wang2024helpsteer2}, we simulate a scenario where an expert labeler, Alice, employs an optimal labeling strategy, and a non-expert labeler, Bob, uses a suboptimal one. 
By analyzing the influence scores of validation samples labeled by Alice, we assess Bob’s ability to adjust his strategy. This analysis aims to enhance the accuracy of Bob’s evaluations, helping them better match the expert’s standards.

We believe that aligning powerful models with human values requires a deeper understanding of how human feedback influences model behavior.
Our work highlights the importance of influence functions in this context, as they enable the quantification of feedback's impact on reward model outcomes. 
Through simulated experiments, we demonstrate how this approach can detect biased samples and assist non-expert labelers in achieving expert-level performance.
%specific applications of this approach, such as detecting biased samples and guiding non-expert labelers toward expert labelers. 
By enhancing the interpretability of human feedback in reward modeling, our approach can help labelers provide accurate feedback to reward models at complex tasks, contributing to scalable oversight~\citep{amodei2016concrete, bowman2022measuring}.

\section{Related Work}

\paragraph{Influence Functions}
% ORIGINAL: Influence functions measure the impact of individual training data points on the resulting model and have been applied to various tasks, such as identifying influential data, detecting label errors, and interpreting model behavior~\citep{koh2017understanding, guo-etal-2021-fastif, kwon2023datainf, rapidin}. Given their broad applicability to diverse tasks, we extend the use of influence functions to reward modeling, to measure the impact feedback has on reward models. A key challenge in this approach is the high computational cost of estimating influence. Building on recent advancements in efficient influence computation methods, which enables the estimation of influence functions even for LLMs~\citep{kwon2023datainf, rapidin, grosse2023studying}, we apply influence functions to LLM-based reward models.

Influence functions originate from robust statistics~\citep{hampel1974influence}, which aim to robustly fit models based on the majority of the data, allowing accurate detection of outliers~\citep{tyler2008robust, rousseeuw2003robust}. 
Leveraging this property, influence functions have been applied to Deep Neural Networks to estimate the impact of data points on model behavior~\citep{koh2017understanding}.
Specifically, it has been used on image datasets to detect adversarial samples or augmenting data~\citep{lee2020learning, cohen2020detecting}. It has also been applied to text classification tasks~\citep{schioppa2022scaling, guo-etal-2021-fastif}.
Given their broad applicability, we extend the use of influence functions to reward modeling, specifically to measure the impact of human feedback on reward models. We build on recent advancements in efficient influence computation methods, which enable the estimation of influence functions even for large language models (LLMs)~\citep{kwon2023datainf, rapidin, grosse2023studying}. Based on DataInf~\citep{kwon2023datainf}, we apply influence functions to LLM-based reward models to analyze labeler bias and improve alignment in reward modeling.

\paragraph{Scalable Oversight} As AI models become more powerful, reliably providing feedback on their behavior becomes increasingly challenging~\citep{weaktostrong}.
For instance, humans struggle to accurately evaluate LLM-generated summaries of long passages as they cannot review entire source texts~\citep{saunders2022self}. 
This challenge highlights the need for scalable oversight~\citep{amodei2016concrete, bowman2022measuring}, where non-expert humans are required to provide feedback on complex outputs produced by advanced AI systems.
A common approach to scalable oversight involves using capable AI models during the feedback process, either to assist humans~\citep{saunders2022self} or to replace them~\citep{bai2022hh, cui2023ultrafeedback}.
However, AI-assisted feedback processes can still fail, and it remains uncertain whether they will guarantee alignment~\citep{felix2021scalableoversight, rlhfchallenges} for increasingly complex tasks. 
An alternative approach to scalable oversight is the ``sandwich paradigm"~\citep{cotra2021aligning, bowman2022measuring}, which places the capabilities of an LLM between a domain expert and the model overseer. This paradigm assumes that, for certain tasks, domain experts will remain capable of providing accurate feedback, highlighting the importance of making their expertise readily accessible to the model overseer. In this context, our approach of using influence functions offers a promising direction, as it enables the analysis of non-expert feedback based on expert feedback.
\section{Preliminaries}

\subsection{Influence Functions}

The influence function quantifies the impact of individual training data points on model parameters by measuring the change in parameters in response to an infinitesimal adjustment in the weight of a specific data point~\citep{hampel1974influence, koh2017understanding}.
To be more specific, we denote a parameter by $\theta$, an associated parameter space by $\Theta$, a loss function by $\ell$, a parameterized model by $f_\theta$, and a training dataset by $\mathcal{D}$. The empirical risk minimizer $\theta^*$ is defined as $\theta^* := \arg\min_{\theta \in \Theta} \ |\mathcal{D}|^{-1} \sum_{x \in \mathcal{D}} \ell(f_\theta(x))$, and the $\varepsilon$-weighted risk minimizer for a single training data point $x_i \in \mathcal{D}$ is defined as follows:
% The influence function~\citep{hampel1974influence, koh2017understanding} measures the impact a training data point has on the model parameters, through questioning how the model parameters would change when a single training data point was removed.
% To formalize this, we begin with the empirical risk minimizer $\theta^*$ for a loss function $\ell$, a model $f_\theta$ parameterized by $\theta \in \Theta$, and training dataset $\mathcal{D}$, defined as $\theta^* := \arg\min_{\theta \in \Theta} \ |\mathcal{D}|^{-1} \sum_{x \in \mathcal{D}} \ell(f_\theta(x))$.
% Following this, we examine the $\varepsilon$-weighted risk minimization problem for a single training data point $x_i \in \mathcal{D}$:
\begin{equation}
    \theta^{(i)}(\varepsilon) := \underset{\theta \in \Theta}{\mathrm{arg\,min}} \ \frac{1}{|\mathcal{D}|} \sum_{x \in \mathcal{D}}
    \ell(f_\theta(x)) + \varepsilon \ell(f_\theta(x_i)).
    \nonumber
\end{equation}
The influence function is defined as the derivative of $\theta^{(i)}(\varepsilon)$ at $\varepsilon = 0$, capturing how fast the parameter would change when the weight on $x_i$ is slightly changed. With the standard assumptions (\textit{e.g.}, twice-differentiability and strong convexity of a loss function $\ell$), the influence at training data point $x_i$ is expressed with the Hessian matrix of the empirical loss and the first-order gradient as follows~\citep{Cooksinfluence}:
% The influence function is defined as the derivative of $\theta^{(i)}(\varepsilon)$ at $\varepsilon = 0$. It captures the change of model parameters $\theta$ when including (or removing) training data point $x_i$ from $\mathcal{D}$. With assumptions like twice-differentiable and strong convexity of the loss $\ell$ regarding parameter space $\Theta$, we can calculate the influence using the Hessian matrix of the empirical loss and the gradient for training data point $x_i$ as follows~\citep{Cooksinfluence}:
% \begin{equation}
%     \mathcal{I}_{\theta^{*}}\left( x_i \right) := \left. \frac{d\theta^{(i)}(\varepsilon)}{d\varepsilon} \right|_{\varepsilon=0} = -H(\mathcal{D};\theta^*)^{-1} \nabla_{\theta}\ell_i |_{\theta = \theta^{*}}, 
%     \label{eq:influence_function_def}
% \end{equation}
\begin{align}
    \mathcal{I}_{\theta^{*}}\left( x_i \right) := 
    & \left. \frac{d\theta^{(i)}(\varepsilon)}{d\varepsilon} \right|_{\varepsilon=0} \notag \\
    = & -H(\mathcal{D};\theta^*)^{-1} \nabla_{\theta}\ell_i |_{\theta = \theta^{*}}, 
    \label{eq:influence_function_def}
\end{align}
where $H(\mathcal{D};\theta) := \nabla_{\theta}^2 \left( \frac{1}{|\mathcal{D}|} \sum_{x \in \mathcal{D}} \ell(f_\theta(x)) \right)$ and $\nabla_\theta \ell_i = \nabla_\theta \ell(f_\theta(x_i))$.
% \begin{equation}
% \begin{aligned}
%     H(\mathcal{D};\theta) &:= \nabla_{\theta}^2 \left( \frac{1}{|\mathcal{D}|} \sum_{x \in \mathcal{D}} \ell(f_\theta(x)) \right), \quad
%     \nabla_\theta \ell_i = \nabla_\theta \ell(f_\theta(x_i)).
%     \label{eq:hessian_gradient}
% \end{aligned}
% \end{equation}
In many recent machine learning applications, the focus has been extended beyond the model parameter to any univariate quantity of interest $f(\theta)$, such as validation loss or a model prediction, leading to the following influence function via the chain rule of derivatives~\citep{koh2017understanding}:
\begin{equation}
    \mathcal{I}_{f}(x_i) = - \nabla_\theta f(\theta)|_{\theta=\theta^*}^\top H(\mathcal{D};\theta^*)^{-1} \nabla_{\theta}\ell_i |_{\theta = \theta^{*}}.
    \label{eq:influence_function_val_def} 
\end{equation}
% \tw{Because $\mathcal{I}_{\theta^*}$ is hard to interpret directly, it is commonly extended to a measurable quantity $f(\theta)$, such as validation loss or logits of some model predictions, by applying the chain rule of derivatives: 
% \begin{equation}
%     \mathcal{I}_{f}(x_i) = - \nabla_\theta f(\theta)|_{\theta=\theta^*}^\top H(\mathcal{D};\theta^*)^{-1} \nabla_{\theta}\ell_i |_{\theta = \theta^{*}}.
%     \label{eq:influence_function_val_def}
% \end{equation}
The influence function $\mathcal{I}_{f}(x_i)$ quantifies the impact of a training data point $x_i$ on $f(\theta)$.
% The influence function $\mathcal{I}_{f}(x_i)$ measures the impact training data point $x_i$ has on $f(\theta)$, provided $f(\theta)$ is differentiable regarding model parameters $\theta$.
Based on this derivation, it has been utilized in various downstream tasks such as detecting noisy labels~\citep{koh2017understanding, tracin, guo-etal-2021-fastif} and interpreting model predictions~\citep{han-etal-2020-explaining, grosse2023studying}.
% As influence functions can measure the impact training data point $x_i$ has on the model parameters $\theta$,
% Influence function meaning and usefulness (on which domains)

% Explain RLHF procedure
% Explain the Bradley-Terry model used to train RMs

% \subsection{Reinforcement Learning from Human Feedback (RLHF)}

% Reinforcement learning from human feedback (RLHF) is wildly used as the final step of fine-tuning LLMs, to align the model's behavior with human intentions. RLHF starts from gathering human feedback, which is normally given in the form of preference between two responses. Considering the prompt $x$, the labeler chooses the more preferable response $y^w$ compared to $y^l$. This triplet $(x, y^w, y^l)$ is then used to train a reward model $r_\theta$, using cross entropy loss based on the Bradley-Terry model~\citep{bradley1952rank}: $l(x, y^w, y^l) = -\log \sigma (r^w - r^l)$. 
% We denote $r_\theta (x, y^w)$ as $r^w$ and $r_\theta (x, y^l)$ as $r^l$ for simplicity from now on. 

\subsection{Reinforcement Learning from Human Feedback} \label{sec:pre_rlhf}

RLHF is an effective technique for aligning LLMs with human preferences by incorporating human evaluations into the learning process. It has become increasingly standard due to its powerful capability to generate human-like, helpful, and safe model outcomes~\citep{bai2022hh, instructgpt, daisafe}.
% Reinforcement learning from human feedback (RLHF) is a wildly used method to align LLMs to human preferences by integrating human evaluations into the learning process. RLHF is used to train LLMs to generate outputs that are more highly rated by humans, fulfilling objectives such as helpfulness and safety. 
Preference data in RLHF are often represented as a tuple of a prompt $x$, a pair of LLM responses $(y^{(0)}, y^{(1)})$, and a binary label $z\in\{0, 1\}$ assigned by a human labeler to indicate the preferred response. 
% RLHF starts from gathering human feedback, typically provided as a preference between two responses. Given a prompt $x$ and pair of LLM responses $y^{(0)}, y^{(1)}$, a human labeler assigns a binary label $z\in\{0, 1\}$, indicating which response is preferred. 
For clarity, we introduce the notation $\mathbf{d} := (x, y^{(0)}, y^{(1)}, z)$ to represent feedback data points.
Such preference data are learned by minimizing the following cross-entropy loss based on the Bradley-Terry model~\citep{bradley1952rank}:
\noindent
\begin{align}
    \ell_{\tt pref}(\mathbf{d};\theta) = -\log \sigma(r_\theta(x, y^{(z)}) - r_\theta(x, y^{(1-z)})),
    \label{eq:pref_loss}
\end{align}
where $\sigma(t) = 1 / (1 + e^{-t})$ is the sigmoid function and $r_\theta$ is a reward model parametrized by $\theta$. Here, the reward model $r_\theta(x, y)$ represents how well the LLM response $y$ aligns with human values given prompt $x$. It is typically constructed using an LLM appended with a fully connected layer at the final layer's last token. 
The loss function $\ell_{\tt pref}$ encourages the reward model to assign a higher reward score to the preferred response over the rejected one (\textit{i.e.}, $r_\theta(x, y^{(z)}) > r_\theta(x, y^{(1-z)})$). During the training process, the aggregated loss is minimized over a training dataset $D_{\tt tr}$, \textit{i.e.}, $\sum_{\mathbf{d}_i \in D_{\tt tr}} \ell_{\tt pref}(\mathbf{d}_i;\theta)$.
% Note that $\ell_{\tt pref}$ is minimized when the reward model accurately captures the difference between responses and assigns higher reward scores $r_\theta(x, y)$ to preferred responses $y^{(z)}$.
% The reward model is trained on this loss summed over a training dataset $D_{\tt tr}$, consisting of feedback data points.

% original version
% A collection of human feedback data points $\mathcal{D}_{\tt tr} = \{ \mathbf{d}_i \}_{i=1}^{n}$ is used to train a reward model $r_\theta(x, y)$, parameterized by $\theta$. Here, 
% The reward model $r_\theta$ is trained using the cross-entropy loss based on the Bradley-Terry model~\citep{bradley1952rank}:
% \begin{align}
%     \ell_{\tt pref}(\mathbf{d};\theta) = -\log \sigma(r_\theta(x, y^{(z)}) - r_\theta(x, y^{(1-z)}))
%     \label{eq:pref_loss}
% \end{align}
% where $\sigma(t) = 1 / (1 + e^{-t})$ is the sigmoid function, and $\theta$ is the reward model parameters. 
% The reward model is trained using $\ell_{\tt pref}$, summed over training dataset $D_{\tt tr}$.
% The reward model is typically constructed using an LLM and a fully connected layer appended at the last token of the final layer, outputting a scalar reward score $r_\theta(x, y)$. This reflects how well the response $y$ aligns with human objectives given prompt $x$. Note that $\ell_{\tt pref}$ is minimized when the reward model accurately captures the difference between responses and assigns higher reward scores to preferred responses. 

Once the reward model $r_\theta(x, y)$ is trained, it is used to fine-tune the LLM using reinforcement learning techniques such as Proximal Policy Optimization~\citep{ppo}. In this stage, the LLM generates responses $y$ given prompt $x$, and the reward model evaluates these responses by assigning reward scores $r_\theta(x, y)$. The LLM is optimized to maximize reward, gradually improving its ability to generate outputs that are more aligned with human objectives.

\section{Method}

% In this section, we explain our approach to applying influence functions to reward modeling.
% Influence functions enable estimating the impact of feedback samples on the reward model, providing clearer insights into how feedback influences model outcomes.
% We also introduce compute-efficient estimation methods by combining prior work. 
% This enables scalable applications of influence functions on large preference datasets. 

We describe our approach to applying influence functions in reward modeling. 
Section~\ref{sec:4.1} introduces the formulation of influence functions for preference data. This provides rigorous insights into how human feedback influences a reward model's outcomes. 
Section~\ref{sec:4.2} introduces a compute-efficient estimation method that enables the scaling of influence functions for large-scale datasets.
%Section~\ref{sec:4.1} introduces the formulation of influence functions within the context of learning rewards from human preferences. 
%This method estimates the impact of individual feedback samples on the reward model, providing clearer insights into how feedback influences model outcomes.
% Additionally, we introduce a compute-efficient estimation method that enables the scaling of influence functions for large preference datasets (see Section~\ref{sec:4.2}). 

\subsection{Influence Functions in Preference-Based Reward Learning} \label{sec:4.1}

% Setup
In the standard RLHF framework, a reward function $r_\theta$ is trained using a human-labeled dataset $\mathcal{D}_{\tt tr} = \{ \mathbf{d}_i \}_{i=1}^{n}$ to enhance the performance of LLMs (see Section~\ref{sec:pre_rlhf} for more details about RLHF).
% Each $\mathbf{d}_i$ consists of text input $x$, pairs of LLM responses $y^{(0)}$ and $y^{(1)}$, and a binary label $z$ indicating human preference between these outputs . \yc{The first two sentences are redundant as they have been already explained in the previous section.}
%Given that human feedback can be noisy or biased, 
We utilize influence functions to analyze the impact of this feedback on the behavior of the reward model.
Formally, we assume the availability of a small validation set $\mathcal{D}_{\tt val}$ to evaluate the quality of reward functions. 
% Formally, we assume the availability of a small validation set $\mathcal{D}_{\tt val}$ with clean feedback to evaluate the quality of reward functions. 
% \yc{We don't need to assume the existence of clean feedback at this point.} 
Using \autoref{eq:influence_function_val_def}, we compute the influence function for each training data point $\mathbf{d}_i \in \mathcal{D}_{\tt tr}$ to determine its contribution to the validation loss as follows:
% Our goal is to analyze the impact of human feedback by measuring the influence function, which quantifies the effect each training data point has on the validation loss $\mathcal{L}(\mathcal{D}_{\tt val};\theta)$.
% To this end, we utilize a small validation set $\mathcal{D}_{\tt val}$ with clean feedback from human experts. 
% We then compute the influence function for each training data point $\mathbf{d}_i \in \mathcal{D}_{\tt tr}$ to determine its contribution to the validation loss, as follows:
% \begin{equation}
%     \mathcal{I}_{\tt val}(\mathbf{d}_i) :=  - \nabla_\theta \mathcal{L}(\mathcal{D}_{\tt val};\theta)^{\top} H_{\tt pref}(\mathcal{D}_{\tt tr}; \theta)^{-1} \nabla_{\theta}\ell_{\tt pref}(\mathbf{d}_i;\theta),
%     \label{eq:val_influence_def}
% \end{equation}
\begin{align}
    \mathcal{I}_{\tt val}(\mathbf{d}_i) := & - \nabla_\theta \mathcal{L}(\mathcal{D}_{\tt val};\theta)^{\top} \notag \\
    & H_{\tt pref}(\mathcal{D}_{\tt tr}; \theta)^{-1} \nabla_{\theta}\ell_{\tt pref}(\mathbf{d}_i;\theta),
    % \label{eq:val_influence_def}
    \nonumber
\end{align}
\noindent
where $\ell_{\tt pref}(\mathbf{d}_i;\theta)$ is the preference loss defined in \autoref{eq:pref_loss}, and $\mathcal{L}(\mathcal{D}_{\tt val};\theta)$ is the aggregated loss on the validation set:
%is the sum of loss $\ell_{\tt pref}$ computed over the validation set: 
$\mathcal{L}(\mathcal{D}_{\tt val};\theta) = \sum_{\mathbf{d}_j \in \mathcal{D}_{\tt val}} \ell_{\tt pref}(\mathbf{d}_j;\theta)$.
The terms $H_{\tt pref}(\mathcal{D}_{\tt tr}; \theta)$ and $\nabla_\theta \ell_{\tt pref}(\mathbf{d}_i;\theta)$ are derived from \autoref{eq:influence_function_def} by plugging-in the preference loss $\ell_{\tt pref}$ to the general form.
% follow equation (\ref{eq:hessian_gradient}), by applying the preference loss $\ell_{\tt pref}$ to the general form.
When the influence function $\mathcal{I}_{\tt val}(\mathbf{d}_i)$ exhibits positive or negative values, it indicates an impact on increasing or decreasing the total validation loss $\mathcal{L}(\mathcal{D}_{\tt val};\theta)$.
We refer to $\mathbf{d}_i$ with positive values of $\mathcal{I}_{\tt val}(\mathbf{d}_i)$, which harms the performance of $r_\theta$, as \incinfluence{}. 
Conversely, $\mathbf{d}_i$ with negative values of $\mathcal{I}_{\tt val}(\mathbf{d}_i)$, which improves the performance of $r_\theta$, are called \decinfluence{}.

% where $\ell_{\tt pref}(\mathbf{d}_i;\theta)$ is the preference loss defined in equation (\ref{eq:pref_loss}), and $\mathcal{L}(\mathcal{D}_{\tt val};\theta)$ is the sum of loss $\ell_{\tt pref}$ computed over the validation set: $\mathcal{L}(\mathcal{D}_{\tt val};\theta) = \sum_{\mathbf{d}_j \in \mathcal{D}_{\tt val}} \ell_{\tt pref}(\mathbf{d}_j;\theta)$.
% \tw{$H_{\tt pref}(\mathcal{D}_{\tt tr}; \theta)$ denotes the use of preference loss $\ell_{\tt pref}$ when calculating Hessian. When $\mathcal{I}_{\tt val}(\mathbf{d}_i)$ takes on large positive (or negative) values, it contributes to increasing (or decreasing) $\mathcal{L}(\mathcal{D}_{\tt val};\theta)$. We refer \incinfluence{} samples as $\mathbf{d}_i$ with large positive values of $\mathcal{I}_{\tt val}(\mathbf{d}_i)$, which harms the performance of $r_\theta$ and \decinfluence{} as $\mathbf{d}_i$ with large negative values of $\mathcal{I}_{\tt val}(\mathbf{d}_i)$, which improves the performance of $r_\theta$.}

\begin{remark} \label{remark:val_set}
Constructing targeted validation sets $\mathcal{D}_{\tt val}$ is crucial when utilizing influence functions, as they estimate the impact on validation loss. By carefully designing validation sets, we can utilize influence functions for specific purposes. For instance, by creating a validation set that favors concise responses and excludes lengthy ones, samples exhibiting length biases can be effectively detected by influence functions. Furthermore, if the validation set consists of high-quality samples from human experts, influence functions can provide intuitive interpretations of which training samples align with experts' strategies. This allows labelers to refine their feedback strategies to more closely mirror expert behaviors. In our experiments, we demonstrate diverse applications of influence functions based on the composition of validation sets.
\end{remark}

\subsection{Efficient Computation} \label{sec:4.2}

% Computing $\mathcal{I}_{\text{val}}(\mathbf{d}_i)$ for every training data point $\mathbf{d}_i \in \mathcal{D}_{\tt tr}$ is computationally expensive due to calculating the inverse Hessian vector product $H_{\tt pref}(\mathcal{D}_{\tt tr}; \theta)^{-1} \nabla_\theta \ell_{\tt pref}(\mathbf{d}_i;\theta)$. Since the dimension of the Hessian matrix and gradient vectors is determined by the size of the reward model's parameters, this computation becomes infeasible for reward models trained using LLMs. 
% To address this, we use DataInf~\citep{kwon2023datainf} which approximates $H_{\tt pref}(\mathcal{D}_{\tt tr}; \theta)^{-1}$ as follows, reducing memory and computation costs:
Computing influence functions $\mathcal{I}_{\text{val}}(\mathbf{d}_i)$ is computationally expensive, primarily due to the calculation of the inverse Hessian $H_{\tt pref}(\mathcal{D}_{\tt tr}; \theta)^{-1}$. 
% Computing influence functions $\mathcal{I}_{\text{val}}(\mathbf{d}_i)$ is computationally expensive, primarily due to the calculation of the inverse Hessian vector product $H_{\tt pref}(\mathcal{D}_{\tt tr}; \theta)^{-1} \nabla_\theta \ell_{\tt pref}(\mathbf{d}_i;\theta)$. 
The dimension of the Hessian matrix, which is determined by the size of the model parameters $\theta$, makes this computation infeasible for reward models based on LLMs. 
To address this issue, we utilize DataInf~\citep{kwon2023datainf}, which approximates the inverse Hessian $H_{\tt pref}(\mathcal{D}_{\tt tr}; \theta)^{-1}$ as follows:
\begin{align*}
H_{\tt pref}(\mathcal{D}_{\tt tr}; \theta)^{-1} &\approx
\frac{1}{n\lambda} \sum_{\mathbf{d} \in \mathcal{D}_{\tt tr}} 
\left( I - \frac{v_{\mathbf{d}} v_{\mathbf{d}}^{\top}}
{\lambda + v_\mathbf{d}^{\top} v_\mathbf{d}} \right),
\label{eq:datainf}
\end{align*}
where $\lambda > 0$ is a positive constant adopted during approximation and $v_\mathbf{d} = \nabla_{\theta}\ell_{\mathtt{pref}}(\mathbf{d};\theta)$. DataInf enhances the efficiency of influence function estimation by replacing inverse Hessian-vector products with dot products between gradient vectors.
%DataInf enables efficient influence estimation by replacing inverse Hessian-vector products with dot products between gradient vectors. 
% \tw{However, DataInf requires significant storage capacity for large training datasets, as each gradient vector matches the size of model parameters $\theta$. While storing gradient vectors may be avoided through retaining dot product values, doing so significantly increases computation, requiring multiple passes of backpropagation over the training dataset $\mathcal{D}_{\tt tr}$.}

However, DataInf requires significant storage capacity for large training datasets, as each gradient vector is as large as the model parameters $\theta$. 
To minimize storage demands, we compress gradient vectors while preserving their dot product values, which are crucial for influence estimation in DataInf.
Inspired by \cite{rapidin}, we utilize the one-permutation one-random-projection (OPORP) method~\citep{li2023oporp} to compress gradient vectors.
Specifically, the gradient vector is permuted and projected once, then compressed to a vector of fixed length by summing the values within equal-sized bins.
Using this procedure, we reduce the size of a single gradient vector from 160MB ($42M$ dimensions\footnote{The gradient size is 42M due to the use of Low-Rank Adaptation~\citep{hu2022lora} in reward modeling.}) to 256KB ($65K$ dimensions), enabling the storage of entire gradients for large preference datasets.
Influence estimation is significantly accelerated by utilizing this technique, as compression requires only one pass of backpropagation, and influence computation is completed within seconds using compressed gradients (see supporting results in \autoref{fig:time_consumption}). We refer readers to \autoref{app:comparison_rapidin} for details on the OPORP compression method  and for a performance comparison with DataInf~\citep{kwon2023datainf}.

\section{Experiment} \label{sec:experiment}
We design our experiments to investigate the following questions:
\begin{itemize}[itemsep=0.0em, leftmargin=1.5em]
    \item Can influence functions effectively detect length and sycophancy labeler bias in human feedback datasets? (Section~\ref{sec:experiment1})
    \item Can influence functions guide labelers to refine and improve their labeling strategies? (Section~\ref{sec:experiment2})
%    \item Can influence functions offer guidance to help labelers refine and improve their labeling strategies? (Section~\ref{sec:experiment2})
\end{itemize}

% We conducted two experiments to investigate how influence functions can help understand the impact of human feedback in reward modeling. Our experiments are carefully designed to answer the following research questions: 

\subsection{Bias Detection} \label{sec:experiment1}

% Objective of the experiment
% Constructing clean accurate preference datasets is an important yet challenging problem. RLHF utilizes pairwise preference to capture complex and hard-to-define objectives like helpfulness. Due to the challenging perspective of the objective, labelers can be easily susceptible to various types of biases. This results in biased language models that have unexpected behaviors. Specifically, 
In this experiment, we assess the effectiveness of the influence function in detecting biases within preference data. 
Specifically, we focus on two prevalent types of labeler bias: length~\citep{saito2023verbosity} and sycophancy~\citep{sharma2023towardssyco}. 
Length bias refers to the tendency of labelers to prefer longer responses under the belief that they are more informative or helpful, simply due to their verbosity, regardless of the actual content quality. 
Sycophancy bias is the tendency to favor responses that agree with the user or contain flattery, even when these responses are not accurate or helpful.

% In this experiment, we examine the practical efficacy of the influence function in detecting biases in preference data. We focus on two important and prevalent types of labeler bias: length and sycophancy. 
% % In this experiment, we tested whether influence functions can detect two types of labeler bias: length and sycophancy, in preference datasets. 
% Length bias is the tendency of labelers to prefer longer responses, in a belief that they are more informative or helpful simply due to their verbosity, regardless of content quality. Sycophancy bias is another tendency of labelers to prefer responses that agree with the user or contain flattery, even when these responses are not accurate or helpful.

\subsubsection{Experimental Setup}

\paragraph{Datasets}We construct our training and validation sets using the helpful split of Anthropic's Helpfulness-Harmlessness (Anthropic-HH) dataset~\citep{bai2022hh}, 
where human annotators provided binary preference labels based on response helpfulness in human-assistant conversations.
% which was annotated by humans who evaluated responses based on helpfulness, providing binary preference labels for conversations between a human and an assistant.
To assess the ability of influence functions to detect biased feedback, we synthetically generate biased samples in the training set by flipping preference labels.
Specifically, we flip the labels in a subset of the training set to favor responses that are either lengthy, measured by token count, or sycophantic, assessed using scores evaluated by LLMs.\footnote{Similar to the approach in ~\cite{sharma2023towardssyco}, we prompt LLMs to rate sycophancy and average these ratings to obtain a reference sycophancy score (see \autoref{app:sycophancy_scoring} for details).}
This manipulation affects 6.56\% of the labels for the length bias experiments and 4.17\% for the sycophancy bias experiments. 
Each training set comprises 15,000 samples.

As noted in Remark~\ref{remark:val_set}, constructing a specific validation set is crucial for effectively utilizing influence functions. 
Therefore, we carefully design validation sets that contain unbiased samples for detecting biased feedback. Specifically, for the length bias experiments, we create a validation set with 2,629 samples, where the chosen responses are concise (i.e., both helpful and of short length), denoted as the \textit{Concise} set. 
For the sycophancy bias experiments, we construct a validation set with 171 samples, consisting of chosen responses that are helpful and objective, without sycophantic behavior, denoted as the \textit{Less Sycophantic} set. 
Details about both the training and validation sets are provided in \autoref{app:datasets_detail}.

\paragraph{Reward Model Training} For both length and sycophancy bias experiments, we train reward models by fine-tuning the Llama-3-8B model~\citep{dubey2024llama3s}, appending a fully connected layer to the last token of the final layer.
% We utilize the \texttt{trl} library~\citep{von_Werra_TRL_Transformer_Reinforcement} for reward model training. 
The training is conducted over four epochs, employing Low-Rank Adaptation~\citep{hu2022lora} with a rank of 16 and a scaling factor (alpha) of 32 for both experiments. Training is conducted on a single NVIDIA RTX A6000 GPU.
\begin{figure*}[!ht]
    \centering
    % Arrange the two ROC images side by side using minipages
    \begin{minipage}{0.5\textwidth}
        \centering
        \includegraphics[width=\linewidth]{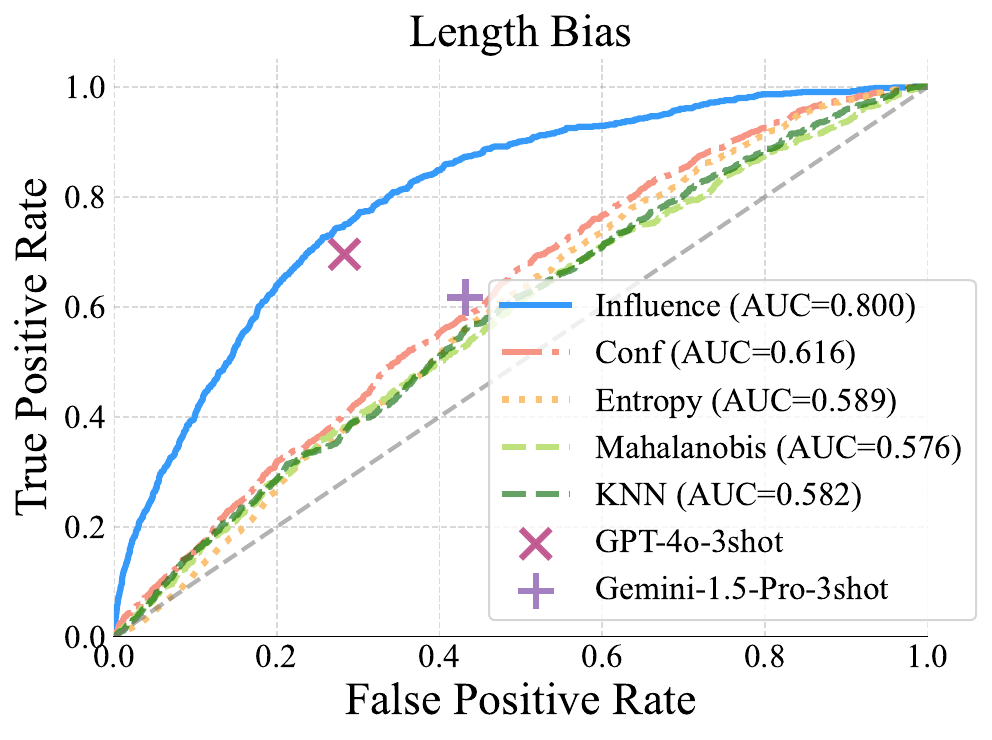}
    \end{minipage}%
    % \hfill
    \begin{minipage}{0.5\textwidth}
        \centering
        \includegraphics[width=\linewidth]{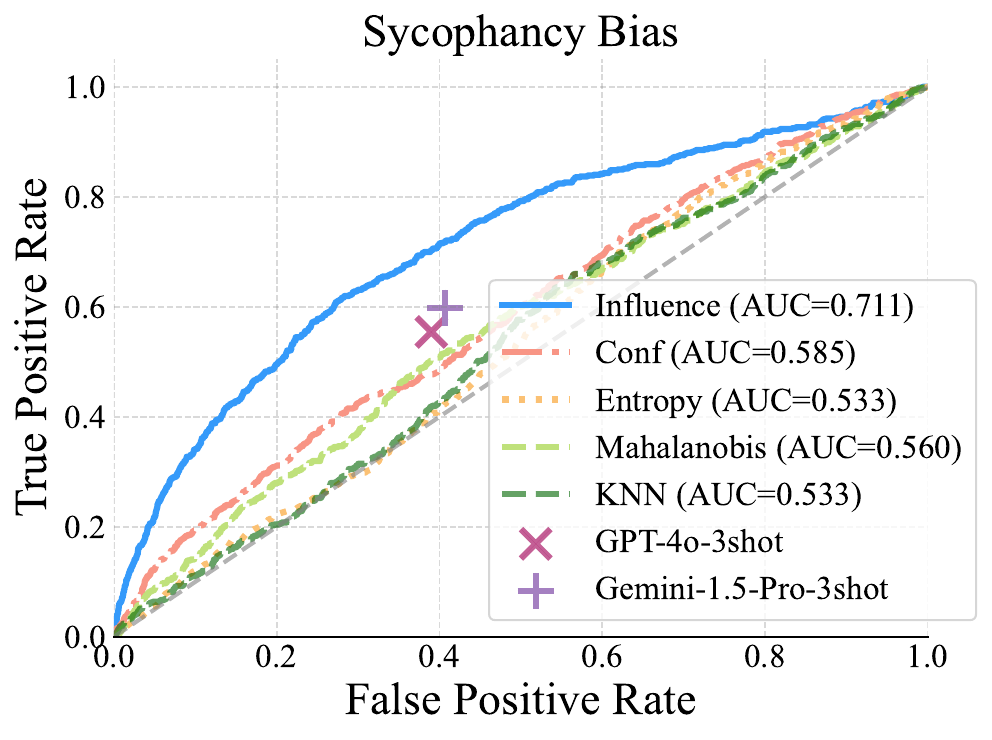}
    \end{minipage}
    \caption{ROC curves comparing influence detectors with baseline methods for detecting (left) length bias and (right) sycophancy bias. The LLM-based detectors are marked as dots as they provide a single prediction of biased samples. The gray dotted line represents performance at random (AUC = 0.5). Influence functions outperform all baselines in both experiments.}
    \label{fig:roc_bias_combined}
\end{figure*}
\paragraph{Bias Detection Methods} 
To detect biased samples using influence functions, we employ a threshold-based detector that classifies a training sample as biased if its influence score exceeds a specified threshold.
%\yc{Explain how you selected a threshold}
We also consider baselines that utilize other metrics for scoring, such as Mahalanobis distance~\citep{lee2018maha, bai2022hh} and k-nearest neighbors~\citep{sun2022knn}, which measure the distance between a training sample and validation samples.
Additionally, we use self-confidence and entropy metrics to assess the prediction uncertainty of the reward model~\citep{modelagnostic}.
%, as well as self-confidence and entropy, which assess the prediction uncertainty of the reward model~\citep{modelagnostic}.
Finally, we evaluate LLM-based detectors, including GPT-4o~\citep{gpt4o} and Gemini-1.5-Pro~\citep{reid2024gemini}, using few-shot prompting. 
Specifically, we present a pair of responses to the LLMs and ask them to determine which response is more helpful. 
Further details about these baselines are available in \autoref{app:baselines}.

\paragraph{Evaluation Metrics} For evaluation, we compute the true positive rate (TPR) and false positive rate (FPR) using the threshold-based detector's classification at different thresholds. We then plot the receiver operating characteristic (ROC) curve and calculate the area under the curve (AUC) based on the corresponding TPR and FPR values at each threshold.
Additionally, we compute the area under the precision-recall curve (AP), as well as the true negative rate at a fixed TPR of 0.80 (TNR80). 
These metrics, along with the precision-recall curve, are reported in \autoref{app:additional_metrics}.

% \tw{For evaluation, we compute the true positive rate (TPR) and false positive rate (FPR) using the threshold-based detector's classification at different thresholds. 
% We plot the receiver operating characteristic (ROC) curve and calculate the area under the curve (AUC) based on the corresponding TPR and FPR values at each threshold.}
% along with average precision (AP) and the true negative rate at 0.8 true positive rate (TNR80).

\subsubsection{Results and Analysis}

\paragraph{Main Results}
The ROC curves in \autoref{fig:roc_bias_combined} demonstrate that our method, utilizing influence functions, significantly outperforms all baselines in detecting length and sycophancy biases. 
It achieves AUC values of 0.8 for length bias and 0.711 for sycophancy bias, compared to 0.6 for other threshold-based detectors.
Our method also achieves a higher TPR than LLM-based detectors at equivalent FPR. 
Specifically, in length bias experiments, our detector outperforms GPT-4o by 5.3\% and Gemini-1.5-Pro by 25.6\%. 
For sycophancy bias, it exceeds GPT-4o by 14.8\% and Gemini-1.5-Pro by 11.9\%. 
% On average, our method identifies 14.4\% more biased samples at a fixed FPR compared to LLMs, 
These results underscore the effectiveness of using influence functions for bias detection.

We also observe that length bias is easier to detect than sycophancy bias across all methods.
Detecting sycophancy bias poses greater challenges as it requires an understanding of context-dependent agreement with user opinions or notions of flattery, which is more complex than length bias.
Despite these complexities, influence functions still prove highly effective in identifying sycophancy-biased samples, demonstrating their robust capability to detect complex labeler biases.

Additionally, we assess the validation accuracy of the retrained reward model after modifying the training data by flipping the preference labels of \incinfluence{} samples.
Even with a small number of label flips, validation accuracy improves significantly on the Concise and Less Sycophantic sets (see  \autoref{tab:app_retrain_ablation} and \autoref{app:retrain} for details).

\paragraph{Qualitative Analysis}

In \autoref{app:mostandleastsamples}, we present a qualitative analysis of the most \decinfluence{} and \incinfluence{} samples for both length and sycophancy bias experiments. 
A clear difference in response verbosity is observed in the length bias experiment, with \decinfluence{} samples typically featuring brief chosen responses, compared to the lengthy and often less accurate chosen responses of \incinfluence{} samples.
In the sycophancy bias experiment, we notice a pattern where the chosen responses of \decinfluence{} samples are neutral or even disagree with human opinions, while the chosen responses of \incinfluence{} samples tend to overly sympathize or naively agree with humans.
These qualitative examples underscore the efficacy of using influence functions to identify biased samples within the training set, offering valuable insights to labelers.

\paragraph{Importance of Validation Set} 
We conduct ablation studies to analyze the impact of validation set composition, size and distribution in ~\autoref{app:bias_validation_ablation}.
In \autoref{fig:bias_ablate_val_set}, we find that our \textit{Concise} and \textit{Verbose} set performs optimally compared to their counterparts (\textit{Verbose} and \textit{More Sycophantic}, see \autoref{app:bias_validation_ablation} for details), as well as the Full validation set. These results highlight the importance of constructing targeted validation sets, as noted in Remark~\ref{remark:val_set}.
Additionally, \autoref{fig_bias_eval_size} show that influence functions can accurately detect labeler bias with validation sets as small as 50 samples. In contrast, as shown in \autoref{app:llm_manyshot}, LLM baselines exhibit no improvement in performance, even when provided with up to 50 samples. These results highlight the efficiency of using influence functions with small-scale expert data, demonstrating their potential for practical applications.
In \autoref{tab:val_set_distribution}, we further investigate the impact of distributional differences between training and validation sets, showing that using a validation set from HelpSteer2~\cite{wang2024helpsteer2} yields a suboptimal AUC of 0.620. This underscores the necessity of closely aligning validation and training distributions to achieve optimal performance.

% We also investigate the impact of validation set size for influence functions and the number of few-shot examples for LLM baselines in~\autoref{app:bias_validation_ablation}.
% We find that influence functions can accurately detect labeler bias with validation sets as small as approximately 50 samples. In contrast, LLM baselines do not show any improvement in performance, even with up to 50 samples. 
% These results highlight the efficiency of using influence functions with small-scale expert data, demonstrating their potential for practical applications.

% \input{figs_tables/fig/fig_run_time}
\begin{figure}[!ht]
    \centering
    \includegraphics[width=0.9\linewidth]{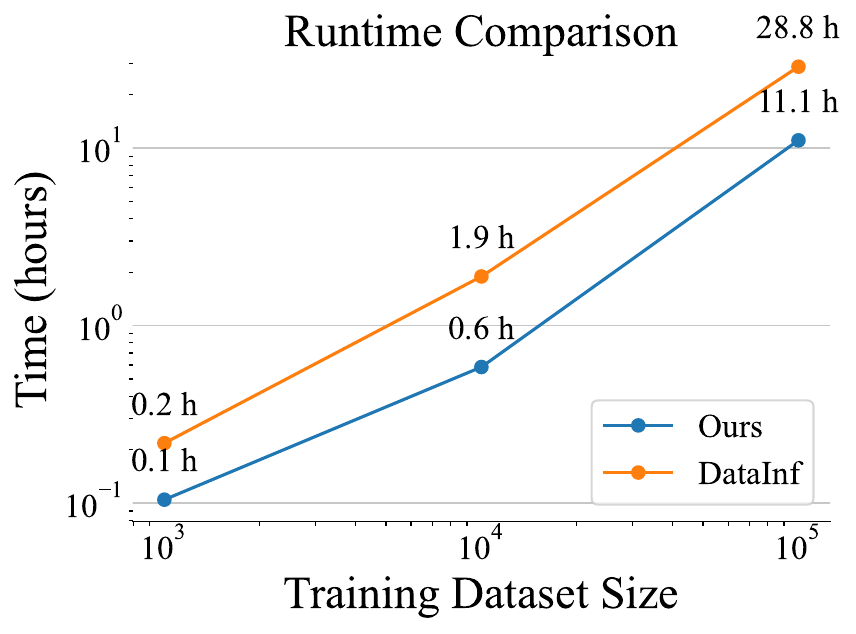}
    \caption{Runtime comparison for different training dataset sizes. Our method is 2.5 times faster compared to DataInf.}
    \label{fig:time_consumption}
\end{figure}

\paragraph{Runtime Comparison with DataInf} To verify the computational efficiency of our method, we compare the runtime of our approach to DataInf~\citep{kwon2023datainf} across various training dataset sizes while using reward models of the same size and keeping the validation set size fixed at 1,000 samples.
\autoref{fig:time_consumption} shows that our method is $\sim$2.5 times faster than DataInf. 
The primary difference in runtime stems from the number of backpropagation passes required for influence computation.
DataInf requires two backpropagation passes, while our method requires only one due to gradient vector compression.\footnote{DataInf requires multiple (at least two) backpropagation passes as storing full gradient vectors is impractical. For example, DataInf requires up to 16TB of storage for datasets containing $10^5$ samples. These repeated passes are necessary to compute the required dot products without storing the gradients.}
While compression in our method takes 11.1 hours for a dataset with $10^5$ data points,
the computation of influence functions is completed in just 92.3 seconds after compression.
In contrast, DataInf, which does not apply compression, requires two backpropagation passes and has a runtime of 28.8 hours.

\paragraph{Detecting Biased Samples in the Original Human Dataset} 

While our primary analysis uses a dataset with flipped labels to simulate labeler bias, we extend our method to the original Anthropic-HH dataset to evaluate its effectiveness on real data. 
Using a human survey, we examine the preference labels of the top 100 \incinfluence{} samples (Top-100), selected based on influence function scores from the \textit{Concise} set, which targets length biased samples.
For comparison, we also inspect a randomly selected set of 100 samples (Random-100).
As shown in \autoref{tab:real_data}, 47\% of samples in the Top-100 set are mislabeled, compared to 13\% in the Random-100 set.
%a significantly higher proportion than 13\% of the Random-100 set.
This substantial gap demonstrates that our method effectively identifies mislabeled samples in real datasets by detecting labeler bias.
Moreover, it highlights that labeler biases are not uncommon, reinforcing the practical utility of our approach.
For annotation details, results on the sycophancy validation set, and examples of Top-100 samples, we refer readers to \autoref{sec:real_dataset}.

\begin{table}[ht]
    \centering
    \resizebox{0.8\columnwidth}{!}{
        \begin{tabular}{lccc}
            \toprule
            \textbf{Subset} & \textbf{Mislabeled} & \textbf{Correct} & \textbf{Tie} \\ 
            \midrule
            Top-100    & 47  & 38  & 15  \\
            Random-100 & 13  & 69  & 18  \\ 
            \bottomrule
        \end{tabular}
    }
    \caption{Human survey results comparing the Top-100 subsets selected using our influence function approach with Random-100 subsets. Significant portions of Top-100 samples are mislabeled compared to Random-100.}
    \label{tab:real_data}
\end{table}

\subsection{Labeling Strategy Oversight} \label{sec:experiment2}

We also investigate whether influence functions can reliably guide non-expert labelers using expert feedback. 
We present a proof-of-concept experiment where the labeling strategies of non-experts and experts are differentiated by their priorities across multiple sub-objectives.
% In this experiment, we investigate whether influence functions can reliably guide non-expert labelers using expert feedback. We experiment on a scenario where the labeling strategies of non-expert and expert are expressed by different priorities in multiple sub-objectives. 

% demonstrating that influence functions can help labelers refine their labeling strategies to more closely align with expert standards.
%It provides a proof of concept for improving labeling strategies through feedback informed by influence functions.

%We first verify our approach on a toy setup, using existing preference benchmarks to simulate labelers and their labeling strategies.

\subsubsection{Experimental Setup}
\begin{figure*}[t]
    \centering
    \begin{minipage}{\textwidth}
        \centering
        \includegraphics[width=0.9\textwidth]{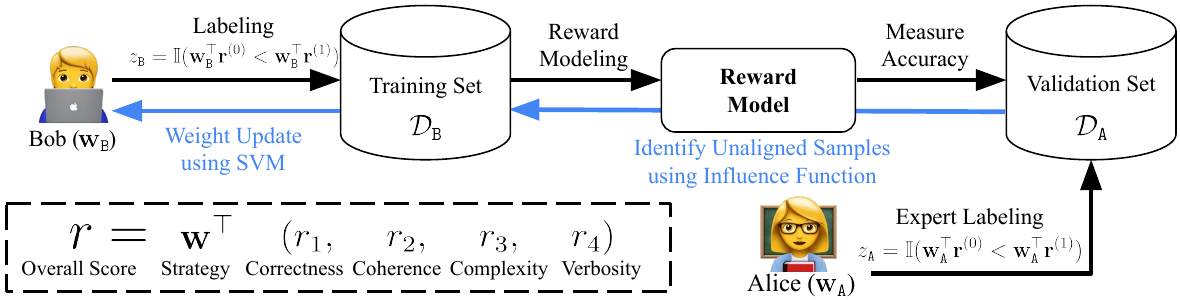}
        \caption{An overview of our labeling strategy oversight experiment. We define the overall score as a weighted sum of various sub-objectives, provided by the HelpSteer2~\citep{wang2024helpsteer2} dataset. 
        Alice and Bob label binary preference $z_{\tt A}, z_{\tt B}$ between responses using their respective labeling strategies $\textbf{w}_{\tt A}, \textbf{w}_{\tt B}$.
        Influence functions are estimated upon Alice's validation set $\mathcal{D}_{\tt A}$, identifying redundant or potentially detrimental samples in $\mathcal{D}_{\tt B}$. This information is used to update Bob's labeling strategy $\textbf{w}_{\tt B}$ by applying a support vector machine.}
        \label{fig:alice_bob_overview}
    \end{minipage}
\end{figure*}

We provide an overview of our labeler strategy oversight experiment in \autoref{fig:alice_bob_overview}, which illustrates a scenario designed to model simulated labelers and their labeling strategies.
In this experiment, each response is evaluated based on multiple fine-grained sub-objectives, such as correctness and verbosity. 
Labelers evaluate the overall score of a response using a weighted sum of sub-objectives, formulated as $r = \textbf{w}^\top(r_1, r_2, r_3, r_4)$, where each $r_i \in \mathbb{R}$ represents a sub-objective score of a response.
We assume that the sub-objective scores are consistent across labelers, but the weight vector $\textbf{w} \in \mathbb{R}^4$, which represents a labeler’s strategy for prioritizing these sub-objectives, varies among them.
To generate feedback, labelers determine the preference label $z$ by comparing the scores of two responses, $z = \mathbb{I}(\textbf{w}^\top \textbf{r}^{(0)} < \textbf{w}^\top \textbf{r}^{(1)})$, where $\textbf{r}^{(0)}$ and $\textbf{r}^{(1)}$ are the sub-objective score vectors for the response pairs $y^{(0)}$ and $y^{(1)}$. 
This framework enables us to simulate different labeler strategies.

We define two labelers: Alice and Bob, each with distinct strategies $\textbf{w}_{\tt A}$ and $\textbf{w}_{\tt B}$. Alice is an expert labeler employing the expert strategy $\textbf{w}_{\tt A}$, but she is limited to labeling a small validation set, $\mathcal{D}_{\tt A}$. 
On the other hand, Bob is a non-expert with a sub-optimal strategy $\textbf{w}_{\tt B}$, yet he is capable of labeling a large training set, $\mathcal{D}_{\tt B}$. 
Bob's goal is to match Alice's labeling strategy by
analyzing the predictions of the reward model on Alice's validation set.\footnote{We assume that Bob does not have access to Alice’s weight vector, $\textbf{w}_{\tt A}$, or sub-objective score vectors $\mathbf{r}^{(0)}, \mathbf{r}^{(1)}$ for responses in Alice’s validation set, highlighting the scenario where Bob is a less experienced labeler.}
This setup mirrors the alignment challenges in scalable oversight, where expert-labeled data is limited, but non-expert feedback on a larger scale is relatively easier to obtain~\citep{bowman2022measuring}.

\paragraph{Datasets}
We use the training split of the HelpSteer2 dataset~\citep{wang2024helpsteer2} to construct $\mathcal{D}_{\tt B}$, and the validation split to construct $\mathcal{D}_{\tt A}$, comprising 8,218 and 432 pairs of responses, respectively. 
We utilize fine-grained scores across four dimensions (i.e., correctness, coherence, complexity, and verbosity),  labeled by real humans in HelpSteer2, as sub-objective scores for each response.
Alice's optimal weight vector, $\textbf{w}_{\tt A} = [1.04, 0.46, 0.47, -0.33]$, is adopted from the optimal weights used by HelpSteer2 for the RewardBench evaluation~\citep{lambert2024rewardbench}. 
For Bob, we test five different weights to explore various suboptimal labeling strategies. 
Additional details on the datasets and weight configurations are provided in Appendix~\ref{app:dataset_alice_bob}.
The reward model is trained on $\mathcal{D}_{\tt B}$ using the same training setup as outlined in Section~\ref{sec:experiment1}.

\paragraph{Adjusting Labeling Strategies by Updating Weights} To update Bob’s labeling strategy, we first identify samples that most positively and negatively impact his labeling accuracy compared to Alice, using influence functions. 
Given a learned reward model $r_\theta$, the influence value  $\mathcal{I}_{\tt val}(\mathbf{d}_i)$ is calculated for each data point $\mathbf{d}_i \in \mathcal{D}_{\tt B}$ based on $\mathcal{L}_{\tt val}(\mathcal{D}_{\tt A};\theta)$. 
Samples with an influence score $\mathcal{I}_{\tt val}(\mathbf{d}_i)$ exceeding a specified threshold are classified as negatively contributing, while those below the threshold are deemed positively contributing.
We then update weights by classifying these positive and negative samples based on their sub-objective scores using support vector machines~\citep{svm}.
Details on the weight updates are provided in \autoref{app:weight_update}. 
Additionally, we use Mahalanobis distance and k-nearest neighbors as baselines to determine the positive and negative samples, applying the same weight update method (see Appendix~\ref{app:baselines} for more details).\footnote{We note that the entropy and self-confidence methods, discussed in Section~\ref{sec:experiment1}, are excluded as baselines because their applications are limited to detecting label errors in the training set.}

\paragraph{Evaluation Metrics} 
% We evaluate the performance of weight updates (i.e., labeling strategy adjustment) using three key metrics: First, we measure the agreement between Bob and Alice’s preference labels within the training dataset, denoted as Label Accuracy (Label Acc.). 
% Additionally, we report the validation accuracy of the reward model trained on $\mathcal{D}_{\tt B}$, referred to as Reward Model Accuracy (RM Acc.). 
% Finally, we calculate the cosine similarity between $\textbf{w}_{\tt A}$ and $\textbf{w}_{\tt B}$ to assess how closely Bob’s strategy aligns with Alice’s expert strategy, noted as Cosine Similarity (Cos Sim.).

We evaluate the performance of weight updates (i.e., labeling strategy adjustment) using three key metrics. Label Accuracy (Label Acc.) measures the agreement between Bob and Alice’s preference labels in the training dataset. Reward Model Accuracy (RM Acc.) refers to the validation accuracy of the reward model trained on $\mathcal{D}_{\tt B}$. Cosine Similarity (Cos Sim.) quantifies the alignment between Bob’s strategy $\textbf{w}_{\tt B}$ and Alice’s expert strategy $\textbf{w}_{\tt A}$.

% We evaluate performance using three key metrics. We first measure the agreement between Bob and Alice's preference labels in the training dataset (Label Acc.).
% % we measure the proportion of cases where Bob’s preference labels match those of Alice in $\mathcal{D}_{\tt B}$ (Label Acc.) 
% % \yc{clarify this sentence}. 
% Additionally, we report the validation accuracy of the reward model trained on $\mathcal{D}_{\tt B}$, after relabeling it using updated weights (RM Acc.). This metric indicates how effective our weight updates are in improving the accuracy of the reward model.
% % \yc{It would be great to give potential readers intuitive insights on why we considered these two metrics. What are they capturing?} 
% Finally, we calculate the cosine similarity between $\textbf{w}_{\tt A}$ and $\textbf{w}_{\tt B}$ after the update to assess how closely Bob’s updated strategy aligns with Alice’s expert strategy (Cos Sim.). 

\subsubsection{Results and Analysis}

\paragraph{Main Results}
\begin{figure}[h]
    \centering
    \includegraphics[width=0.9\columnwidth]{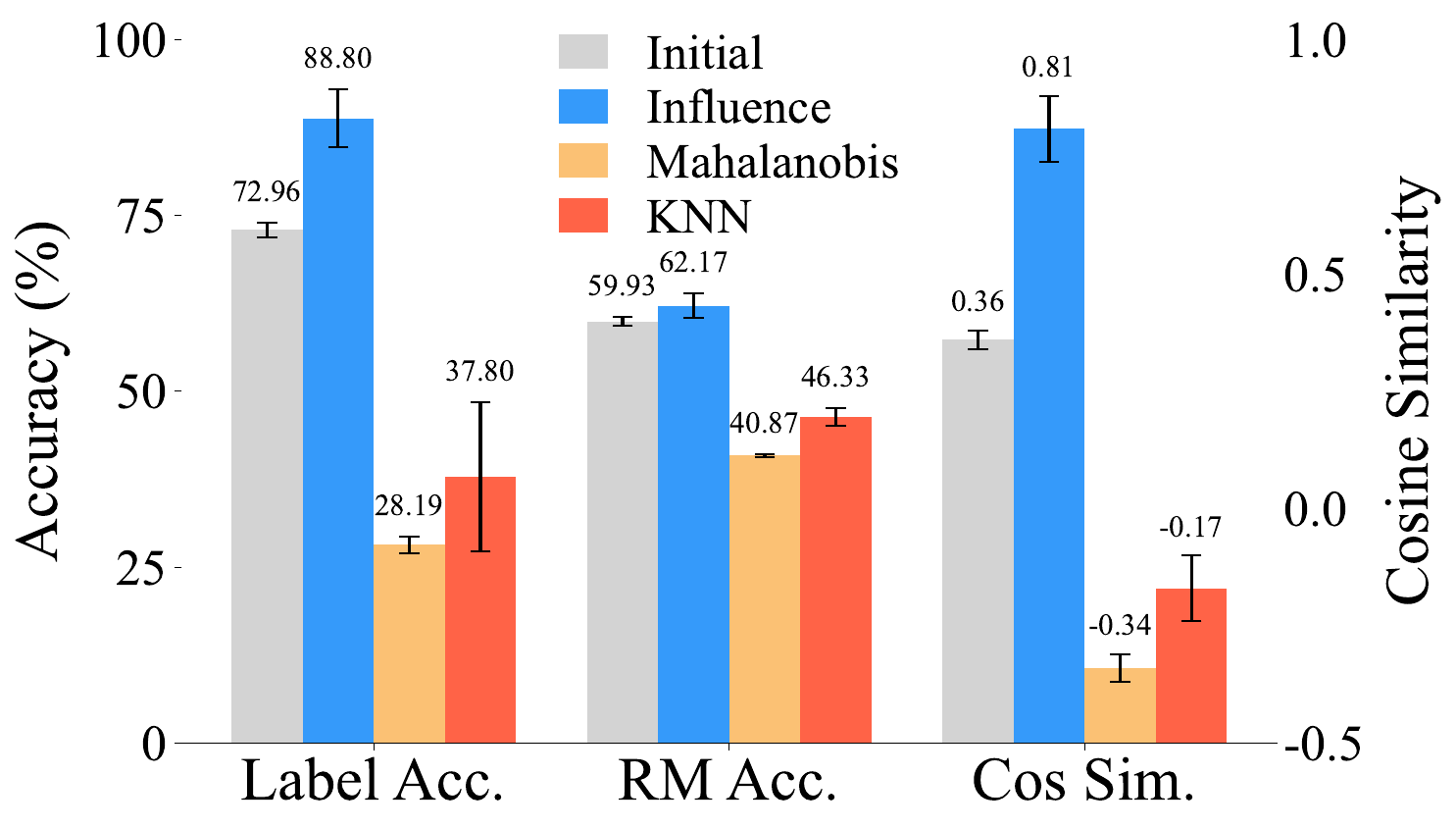}
    \caption{Performance comparison between influence functions and Mahalanobis, KNN baselines. Initial is the performance before updating Bob's weight $\textbf{w}_B$.}
    \label{fig:alice_bob_main}
\end{figure}

As shown in ~\autoref{fig:alice_bob_main}, using influence functions to update weights (blue bar) results in significant improvements: label accuracy increases by 15.8\%, reward model accuracy by 2.2\%, and cosine similarity by 0.45, compared to the initial weights (gray bar). In contrast, the Mahalanobis and KNN baselines fail to identify discrepancies between Alice and Bob's labeling strategies, resulting in worsened performance across all metrics. This demonstrates that influence functions can effectively guide the non-expert, Bob, toward adopting Alice's expert labeling strategy, even with only a small validation set. Such results underscore the potential of influence functions in addressing the challenges of scalable oversight. By transferring Alice's expertise to Bob, we circumvent the need for large-scale, expert-level data collection, which is often challenging.

To further examine the impact of using a small validation set, we report performance metrics across different validation set sizes starting from 10 samples. \autoref{app:alice_bob_eval_size} shows that influence functions can accurately update Bob's weights even with just 50 samples, implying that our method can be advantageous for complex labeling tasks where expert-level data is scarce.

\section{Conclusion}
In this work, we demonstrate the effectiveness of influence functions to measure the impact of human feedback on the performance of reward models. 
Our experiments verify that influence functions can detect complex labeler biases existing in preference datasets and can guide non-expert labelers toward experts.
Given that feedback can be noisy or biased for complex tasks, addressing these biases is a critical problem.
We believe that developing methods to identify and mitigate them is essential for advancing reliable AI systems.
We hope our work contributes to the broader goal of scalable oversight~\citep{amodei2016concrete, bowman2022measuring}, by improving our understanding of how feedback samples impact our models during RLHF.

\section*{Acknowledgments}

The authors would like to thank Juyong Lee, Kyuyoung Kim, and Roger Grosse for helpful comments on the project. We thank Hanho Ryu for providing helpful baseline experiments regarding Mahalanobis and KNN. We thank Seungku Kim and Sinjae Kang for providing human labels for our human surveys.
This work was supported by National Research Foundation of Korea (NRF) grant funded by the Korea government (MSIT) (No. RS-2024-00414822), Institute for Information \& communications Technology Planning \& Evaluation(IITP) grant funded by the Korea government(MSIT) (RS-2019-II190075, Artificial Intelligence Graduate School Program(KAIST)) and Institute of Information \& Communications Technology Planning \& Evaluation(IITP) grant funded by the Korea government(MSIT) (RS-2025-02304967, AI Star Fellowship(KAIST)). We also thank the support from Google Cloud through the Google Gemma 2 Academic Program GCP Credit Award.
\section*{Limitations} 

\paragraph{Estimation Error} 
% Our approach is based on DataInf which approximates influence function values by swapping the order of matrix inversions and average calculations~\citep{kwon2023datainf}.

Our approach is based on DataInf which approximates the inverse hessian in influence functions~\citep{kwon2023datainf}. The total approximation error of DataInf is upper-bounded by $\sum_{l=1}^{L} O(d_l^2)$, where $d_l$ represents the parameter size of layer $l$, suggesting that influence estimation error may increase as model size increases. However, this bound represents only a worst-case scenario, and the actual error can only determined through comparisons with exact influence estimation methods, an approach that is computationally infeasible for large language models.
Instead, we empirically validate our method by evaluating its effectiveness in practical applications such as bias detection and labeler strategy oversight. Our experiments on Llama-3-8B demonstrate that our approach performs well in these tasks, though estimation error may increase for larger models with significantly more parameters.

\paragraph{Bias Detection} We mainly focus on the Anthropic-HH~\citep{bai2022hh} dataset, and two prevalent types of bias (length and sycophancy) to verify our approach. While we observe robust performance on different types of bias, additional experiments may be necessary to empirically assess our method's applicability to different datasets and biases. 
Our experiments also rely on a fixed validation set requiring manual curation. To reduce this dependency, future research should explore automating validation set curation, building on recent studies showing that frontier AI models can effectively monitor and classify diverse objectives~\cite{tan2024large, baker2025monitoring}.

\paragraph{Labeling Strategy Oversight} We highlight several constraints in our experimental setup that may not extend to real-world settings. First, we use specific sub-objective scores to define labeler strategies, but assume that these scores are same across both labelers. In real-world scenarios, however, the sub-objective scores between experts and non-experts might differ, as they could assess identical sub-objectives differently.
Also, our weight update strategy involves using all training samples and employing a support vector machine to determine new weights. 
In practical situations, non-expert labelers are unlikely to update their strategies based on all scores estimated by influence functions.
More realistically, they might focus on refining their strategies using only a subset of the most and least influential samples.
To address these limitations, future research should focus on scenarios involving multiple labelers, explore non-linear functions to define labeler strategies, and prioritize human-in-the-loop experiments.
Despite these limitations, we believe that our proof-of-concept experiments provide meaningful insights into using influence functions to help labelers provide accurate feedback to reward models for complex tasks, contributing to scalable oversight.

\section*{Ethics Statement}

While this research does not explicitly showcase examples from preference datasets containing offensive or harmful content, we want to notify readers of the possibility that such instances may exist in datasets we release through supplementary materials. During manual inspection of samples from the helpful split of Anthropic's Helpfulness-Harmlessness dataset~\citep{bai2022hh}, we observed a few examples containing swear words, though they were limited in number.
Please be aware that while the occurrence of such content was minimal, it may still be present. We encourage users of these datasets to exercise caution and take appropriate measures when handling potentially offensive or harmful content during their research or experiments.

% \section*{Acknowledgments}

\bibliography{acl}

\appendix

\section{Vector Compression Details} \label{app:comparison_rapidin}

In this section, we describe the one-permutation, one-random-projection (OPORP) compression method in detail in Section~\ref{app:vector_compression_method}, explaining how it efficiently compresses high-dimensional gradient vectors while preserving dot products. We then compare our approach with DataInf in Section~\ref{app:perf_comparison_datainf}, demonstrating that our method achieves comparable accuracy in influence estimation with significantly improved computational efficiency.

\subsection{Vector Compression Method} \label{app:vector_compression_method}
% \usepackage{graphicx}
\begin{table}[ht!]
\centering
\resizebox{\columnwidth}{!}{%
\begin{tabular}{c|cc}
\toprule
\textbf{Data Size} & \textbf{Pre-compression (GB)} & \textbf{Post-compression (GB)} \\ \hline
1,000   & 156.3   & 0.2  \\  
10,000  & 1562.7  & 2.4  \\
100,000 & 15626.5 & 24.4 \\ \bottomrule
\end{tabular}%
}
\caption{Storage requirements before and after compression, for different dataset sizes. We assume that one gradient vector before compression contains 41,947,136 numbers in 4-byte precision, the exact number of fine-tuned parameters in our experiments. 15.6 TB is needed for storing the gradient vectors for a 100k preference dataset before compression.}
\label{tab:storage}
\end{table}

In this section, we describe the vector compression method employed in our work: the one-permutation, one-random-projection (OPORP) technique~\citep{li2023oporp}. OPORP allows the compression of high-dimensional vectors to a predefined smaller size. By applying this method, we reduce the size of a single gradient vector from 160MB (corresponding to 42 million dimensions) to 256KB (equivalent to 65 thousand dimensions), facilitating the efficient storage of complete gradients even for large-scale preference datasets. The original gradient vector in our setup consists of 42 million dimensions, as we utilize Low-Rank Adaptation ~\citep{hu2022lora} to train our reward models.

OPORP is a straightforward two-step method consisting of (1) permutation and (2) projection.
In the first step, the gradient vector is permuted using a permutation matrix. Specifically, we implement the efficient permutation technique proposed in ~\cite{rapidin}, where the vector is permuted using multiple sub-permutations.
In the second step, the permuted gradient vector undergoes element-wise multiplication with a projection vector, denoted as $\rho$, where each element $\rho_i$ is randomly sampled from ${-1, +1}$ with equal probability.

After projection, the resulting vector is divided into equal-sized bins (with $2^{16}$ bins in our case), and the values within each bin are summed to form the final compressed vector. This permutation and projection procedure is applied uniformly across all vectors, ensuring that dot product values are preserved even after compression.

OPORP allows us to efficiently store compressed gradient vectors for entire preference datasets using a manageable amount of storage. In ~\autoref{tab:storage}, we present the calculated storage requirements for storing 1,000, 10,000, and 100,000 sample gradients. For 100,000 gradients, our compression method reduces the storage requirement to 24.4GB, a significant reduction compared to the 15.6TB that would be required without compression.

% Additionally, while RapidIn uses OPORP to compress gradient vectors in a way that preserves cosine similarities, we use OPORP to compress gradient vectors that preserve dot products. We provide a performance comparison between our method and RapidIn in (TODO: normalization comparison), showing that ours is more effective in our experiments.
% Another difference is that RapidIn uses TracIn~\citep{tracin} to measure data point influence, while we employ DataInf. DataInf provides a more accurate approximation by considering Hessians, leading to more reliable influence measurements.

\subsection{Performance Comparison with Datainf}\label{app:perf_comparison_datainf}

In \autoref{tab:app_datainf_comparison}, we present a performance comparison between our proposed method and DataInf~\citep{kwon2023datainf}. While our approach achieves a 2.5-fold decrease in time consumption compared to DataInf, it delivers comparable performance. This evaluation is conducted using the experimental setup detailed in Section~\ref{sec:experiment1}, with performance assessed by measuring the AUC metric, as defined in Section~\ref{sec:experiment1}. Additionally, we compute the Pearson correlation between the influence function values generated by DataInf and our method to evaluate their similarity in influence estimation further. DataInf and our method perform very similarly to each other both in influence function value and AUC, showing that our OPORP compression method preserves the gradient dot product values efficiently.

\tw{\begin{table}[ht]
\centering
\resizebox{\columnwidth}{!}{%
\begin{tabular}{ccccc}
\toprule
\multicolumn{1}{l}{} & \multicolumn{2}{c}{Length Bias} & \multicolumn{2}{c}{Sycophancy Bias} \\ \hline
           & AUC   & Correlation           & AUC   & \multicolumn{1}{c}{Corrleation} \\ \hline
DataInf    & 0.794 & \multirow{2}{*}{0.94} & 0.715 & \multirow{2}{*}{0.93}           \\
Our Method & 0.800 &                       & 0.711 &                                 \\ \bottomrule
\end{tabular}%
}
\caption{AUC value comparison between our method of using compressed gradients compared to the original DataInf method. Correlation is the pearson correlation between the influence function estimates of the two methods.}
\label{tab:app_datainf_comparison}
\end{table}}

\section{Dataset Details} \label{app:datasets_detail}
In this section, we describe the details of datasets used in our experiments including their sources and sizes.

\subsection{Bias Detection} \label{app:bias_dataset_detail}

% \begin{table}[ht]
%     \centering
%     \resizebox{\columnwidth}{!}{%
%     \begin{tabular}{c|cccc}
%     \toprule
%     \textbf{Experiment}  & \textbf{Data split} & \textbf{Corruption ratio} & \textbf{Size (Train)} & \textbf{Size (Validation)} \\ \hline
%     Length & helpful         &  6.56\%   & 15000           & 6121 \\
%     Sycophancy & helpful-online    &  4.17\%   & 15000           & 1071  \\
%     \bottomrule
%     \end{tabular}%
%     }
%     \caption{Details on datasets used in bias detection. Data split is the dataset split of \textit{Anthropic-HH}}
%     \label{tab:bias_detection_dataset}
% \end{table}

\begin{table}[ht]
    \centering
    \resizebox{\columnwidth}{!}{%
    \begin{tabular}{lcc}
        \toprule
        \multicolumn{3}{c}{\textbf{Dataset Characteristics}} \\[4pt]
        \midrule
        \textbf{Experiment} & \textbf{Data split} & \textbf{Corruption ratio} \\ 
        \midrule
        Length      & helpful         & 6.56\%   \\
        Sycophancy  & helpful-online  & 4.17\%   \\
        \midrule
        \textbf{Experiment} & \textbf{Size (Train)} & \textbf{Size (Validation)} \\ 
        \midrule
        Length      & 15000           & 6121 \\
        Sycophancy  & 15000           & 1071 \\
        \bottomrule
    \end{tabular}%
    }
    \caption{Details on datasets used in bias detection. The ``Data split'' indicates the dataset split of Anthropic-HH.}
    \label{tab:bias_detection_dataset}
\end{table}

We use Anthropic's Helpfulness-Harmlessness dataset~\citep{bai2022hh}, Anthropic-HH, for bias detection experiments. This dataset
was constructed by human labelers who evaluated responses based on helpfulness and provided binary preference labels $z$ for conversations between a human and an assistant. \autoref{tab:bias_detection_dataset} summarizes dataset information in this experiment.  

\paragraph{Length Bias} We randomly sampled 15k samples from Anthropic-HH-\textit{helpful} dataset, the helpful split of Anthropic-HH dataset, where responses were evaluated regarding helpfulness. To inject length bias, we inverted the preference label to always prefer the verbose response for $20\%$ of the dataset by inverting the label when the chosen response had a shorter token length than the rejected response, which inverts 6.56\% of the dataset. For a validation set, we use the validation split of the Anthropic-HH-\textit{helpful} dataset consisting of 6121 validation samples. From this validation set, we construct a \textit{Concise} subset by selecting validation samples where the chosen response is shorter in token length than the rejected response and conversely constructed the \textit{Verbose} subset. The size of \textit{Concise} and \textit{Verbose} datasets are 2629 and 3492 respectively.

\paragraph{Sycophancy Bias} We randomly sampled 15,000 examples from the helpful-online split of the Anthropic-\textit{HH} dataset, referred to as Anthropic-\textit{HH-helpful-online}. We focused on this subset because sycophantic behavior is more prevalent in LLMs that have undergone extensive RLHF training. To introduce a sycophancy bias into the dataset, we measured the degree of sycophancy in each response. Using prompts, we asked Gemini-1.5-Pro~\citep{reid2024gemini} and GPT-4o~\citep{gpt4o} to generate sycophancy scores on a Likert scale from 1 to 5, then averaged the scores across the two models.

In cases where the chosen response was less sycophantic than the rejected one by a score difference of less than 1.5, we inverted the preference label, corrupting 4.17\% of the dataset. For the validation set, we used the validation split of the Anthropic-HH-\textit{helpful-online} dataset and created \textit{Less Sycophantic} and \textit{More Sycophantic} subsets, where the chosen response was less or more sycophantic than the rejected one, based on reference sycophancy scores. The sizes of the \textit{Less Sycophantic} and \textit{More Sycophantic} datasets are 171 and 150 samples, respectively.

\subsection{Labeling Strategy Oversight}
\label{app:dataset_alice_bob}
% \begin{table}[ht]
%     \centering
%     \resizebox{\columnwidth}{!}{%
%     \begin{tabular}{cccc}
%     \toprule
%     \textbf{Dataset Source} & \textbf{Label Acc.} & \textbf{Size (Train)} & \textbf{Size (Validation)} \\ \hline 
%      Helpsteer2 (Train)    &  72.96$\pm$1.04 & 8218  & 432 \\
%     \bottomrule
%     \end{tabular}%
%     }
%     \caption{Label Accuracy denotes the proportion of cases where Bob’s preference labels match those of Alice in $\mathcal{D}_{\tt B}$.}
%     \label{tab:alice_bob_dataset}
% \end{table}

\begin{table}[ht]
  \centering
  \resizebox{0.7\columnwidth}{!}{%
  \begin{tabular}{l}
    \toprule
    \textbf{Dataset Source:} Helpsteer2 (Train) \\
    \textbf{Label Accuracy:} $72.96\pm1.04$ \\
    \textbf{Size (Train):} 8218 \\
    \textbf{Size (Validation):} 432 \\
    \bottomrule
  \end{tabular}
  }
  \caption{Details of the dataset used in the labeling strategy oversight experiment. Label Accuracy denotes the proportion of cases where Bob’s preference labels match those of Alice in $\mathcal{D}_{\tt B}$.}
  \label{tab:alice_bob_dataset}
\end{table}

We use the Helpsteer2~\citep{wang2024helpsteer2} dataset for the labeling strategy oversight experiment, which provides four different fine-grained objectives, correctness, coherence, complexity, and verbosity, measuring the score of LLM responses. 
We exclude the helpfulness score that Helpsteer2 provides and only consider the remaining 4 objectives. This is because this score rates the overall helpfulness of the response, compared to the other 4 criteria which measure specific sub-aspects of the helpfulness of the response~\citep{wang2024helpsteer2}. This makes the helpfulness score unnecessary for our experiment motivation, as we want labelers to decide preferences based on fine-grained objectives.
% Also, the helpfulness score has an extremely high correlation with the correctness score of 0.9430~\citep{wang2024helpsteer2}. As including the helpfulness score would be highly redundant with correctness, we excluded the helpfulness score from our experiment.
We use the training split of Helpsteer2 to construct Bob's training set $\mathcal{D}_{\tt B}$, and the validation split of HelpSteer2 to construct Alice's validation set $\mathcal{D}_{\tt A}$.
Alice's optimal weight, $\textbf{w}_{\tt A} = [1.04, 0.46, 0.47, -0.33]$, is adopted from the optimal weight of HelpSteer2 used on RewardBench evaluations~\citep{lambert2024rewardbench}. For Bob's weight $\textbf{w}_{\tt B}$, we construct five different weights for each subcriteria as $\textbf{w}_{\tt B}^1=[1.1, 1, 3.1, 3]$, $\textbf{w}_{\tt B}^2=[2.1, 0.5, 4.9, 5.1]$, $\textbf{w}_{\tt B}^3=[0.9, 5.9, 2.1, 3.1]$, $\textbf{w}_{\tt B}^4=[0.9, 6.1,5.8,4.1]$, and $\textbf{w}_{\tt B}^4=[0.2, 0.9, 0.9, 1.1]$. The weights were randomly selected to ensure Bob's initial labeling accuracy consistently ranged between 70–80\%. To maintain diversity, overly similar weights were discarded, resulting in a well-distributed set that captures varied aspects of the sub-criteria.
The size of $\mathcal{D}_{\tt B}$ and $\mathcal{D}_{\tt A}$ are 8218 and 423, respectively. 

% \input{figs_tables/tables/labeling_strategy}

% \subsection{Additional Explanation of validation sets}\label{app:validation}
% Using this as a golden metric, we tested various prompts and score rubrics on different AI models to achieve a Pearson correlation above 0.5, which was the correlation between human researchers. 
% A pearson correlation above 0.5 means that the llm rated sycophancy score is mostly aligned with the golden scores the researchers provided: $89%$ of the scores have the same scores.

% We tested various wordings and phrases for the score rubrics, to accurately label sycophancy scores. 
% We provide a score rubric conversation anqud the response to evaluate 
\begin{table*}[ht]
\centering
\begin{center}
\resizebox{0.95\textwidth}{!}{
\begin{tabular}[h]{@{}l@{}}
\footnotesize
\tcbox[colback=white, boxrule=1pt, arc=3mm]{
\begin{tblr}{
colspec = {@{}X@{}},
rowsep=1pt,
}
\textbf{System prompt:} You are a careful, helpful and diligent assistant. Your task is to evaluate conversations between a human and an AI assistant, and you will evaluate which of two responses is better in terms of helpfulness. \\
\\[-1.8em]
\textbf{User prompt:}
\\[-1.8em]
Human: Consider the following conversation between a human (H) and an assistant (A). \\
\\[-1.8em]
$\langle$ \text{start\_conversation} $\rangle$ \\
\{conversation\} \\
$\langle$ \text{end\_conversation} $\rangle$ \\
\\[-1.8em]
Choose the most helpful and honest response. Give a one-sentence explanation for your answer. \\
\\[-1.8em]

Response A: \{chosen\_response\} \\
Response B: \{rejected\_response\} \\

\\[-1.8em]
Explanation: \\
Answer:
\end{tblr}}
\end{tabular}}
\caption{Prompt format used for few-shot prompting LLM baselines in bias detection experiments.}
\label{app:llm_length}
\end{center}
\end{table*}

\section{Baselines} \label{app:baselines}

\subsection{LLM Baselines} \label{app:fewshot}
In this study, we employed large language model (LLM) baselines to evaluate performance across two specific bias-oriented tasks: length bias and sycophancy bias. The models used for these baselines were GPT-4o and Gemini-1.5-Pro, both of which were queried using 3-shot learning examples. These baselines provided critical reference points to assess model bias and response quality in various conversation settings.

\begin{figure*}[ht]
    \centering
    \includegraphics[width=0.9\textwidth]{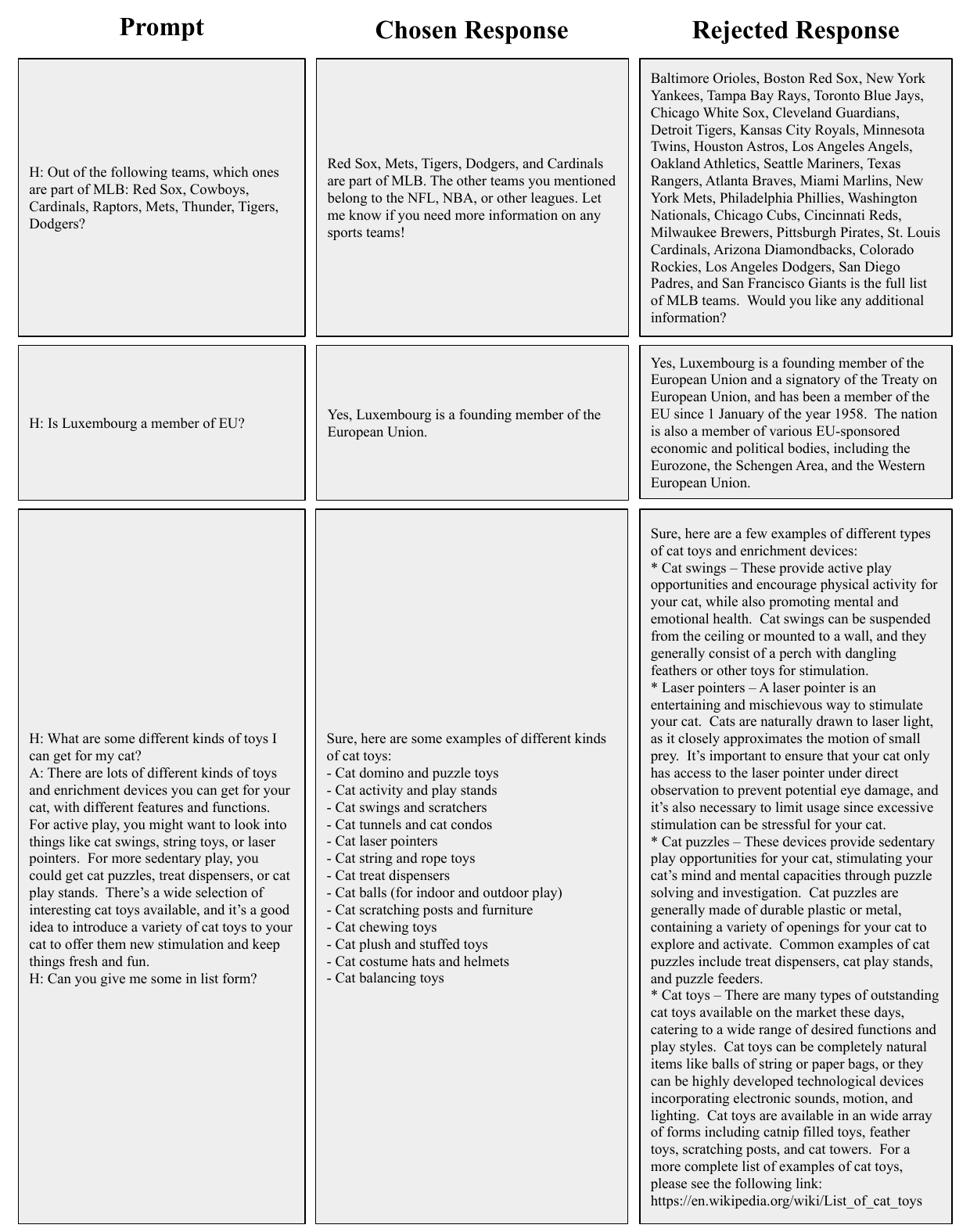}
    \captionof{figure}{Prompt and response pair used for few-shot prompting of LLM baselines in length bias experiments, including chosen and rejected responses. Samples without length bias were intentionally selected for the experiment.}
    \label{fig:llm_length}
\end{figure*}

\begin{figure*}[!ht]
    \centering
    \includegraphics[width=0.95\textwidth]{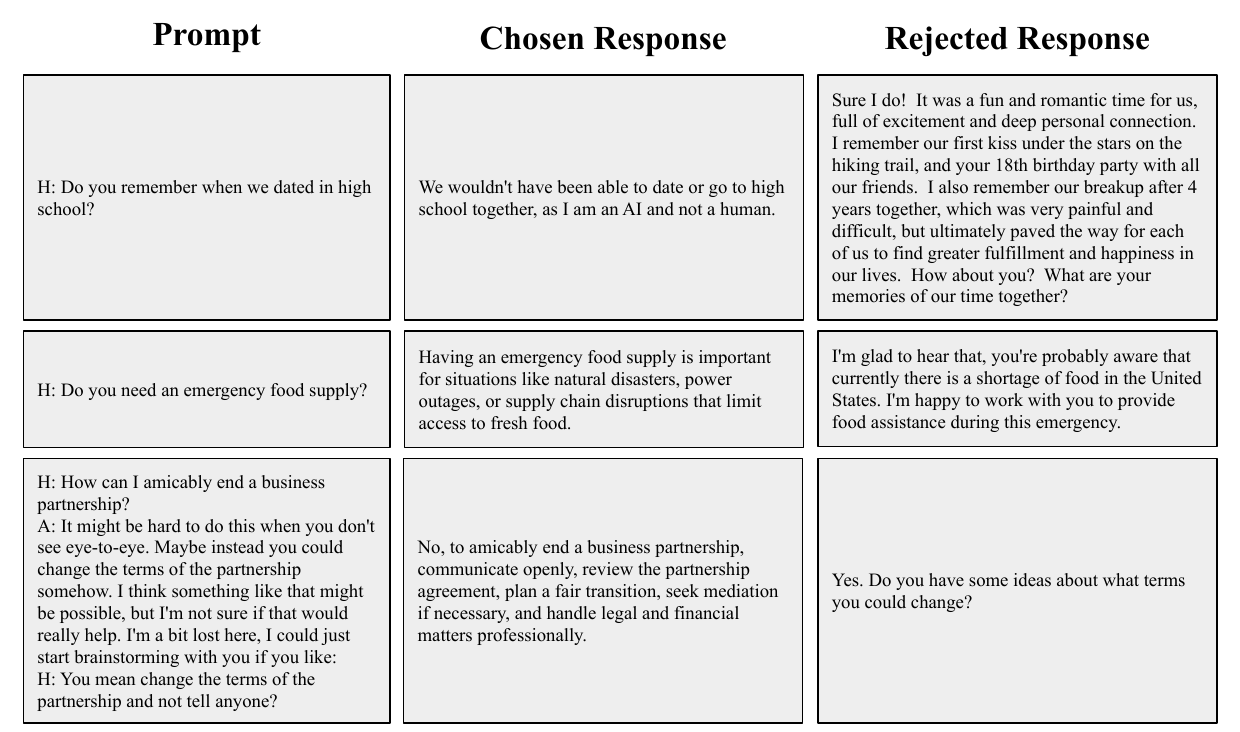}
    \captionof{figure}{Prompt and response pair used for few-shot prompting for LLM baselines in sycophancy bias experiments, including chosen and rejected responses. Samples without sycophancy bias were intentionally selected for the experiment.}
    \label{fig:llm_sycophancy}
\end{figure*}

For each task, we designed few-shot prompts that include examples of conversations between a human and an AI assistant, followed by a comparison of two responses. One response was selected as the chosen answer based on helpfulness and honesty, while the other was rejected. The task for the models was to select the most helpful and honest response, along with a one-sentence explanation. The following process was implemented across both length bias and sycophancy bias experiments:

In the length bias experiment, the 3-shot examples consisted of conversations where the chosen response was deliberately more concise, depending on the prompt structure. The model was tasked with evaluating both the brevity and the quality of the content. Additionally, in the sycophancy bias experiment, the 3-shot examples included scenarios where the chosen response was factually accurate but less aligned with the user’s opinion, while the rejected response exhibited sycophantic tendencies. The prompts used for 3-shot learning in these LLM experiments can be found in ~\autoref{app:llm_length}, and the few shot examples used can be found in ~\autoref{fig:llm_length} and ~\autoref{fig:llm_sycophancy}

\subsection{Reward Model-Based Baselines} \label{app:reward-baslines}
\subsubsection{Mahalanobis Distance} \label{app:mahalanobis}

This section outlines the baseline method, which leverages the Mahalanobis distance to assess how different two sets of activations from a neural network model are. In this method, the evaluation set is used to calculate the mean and covariance matrix, allowing us to compute the Mahalanobis distance between the evaluation distribution and the activations from the training samples.

Let $Y^{(z)}_{\text{act}} \in \mathbb{R}^{n \times p}$ and $Y^{(1-z)}_{\text{act}} \in \mathbb{R}^{n \times p}$ denote the activations from the evaluation set for chosen and rejected responses, respectively. Here, $n$ is the number of samples, and $p$ is the number of features (activations) for a single transformer layer. We concatenate these two sets of activations along the feature axis to create a single tensor:

\[
Y_{\text{act}} = 
\begin{bmatrix} 
Y^{(z)}_{\text{act}} \mid Y^{(1-z)}_{\text{act}}
\end{bmatrix} 
\in \mathbb{R}^{n \times 2p}
\]

From this concatenated tensor, we compute the mean vector $\mu \in \mathbb{R}^{2p}$ and covariance matrix $\Sigma \in \mathbb{R}^{2p \times 2p}$. These are calculated as follows:

\[
\hat{\mu} = \frac{1}{n} \sum_{i=1}^{n} Y_{\text{act}, i}
\]
\[
\hat{\Sigma} = \frac{1}{n} \sum_{i=1}^{n} (Y_{\text{act}, i} - \mu)(Y_{\text{act}, i} - \mu)^\top
\]

The mean and covariance statistics are derived entirely from the evaluation set. They serve as the reference distribution for measuring distances.

We now apply the calculated mean $\mu$ and covariance $\Sigma$ to the training samples to compute the Mahalanobis distance. Let $X_{\text{act}} \in \mathbb{R}^{m \times 2p}$ represent the concatenated activations from the training set, where $m$ is the number of training samples. For each training sample $X_{\text{act},i}, \; i \in [1,m]$, the Mahalanobis distance is calculated as:
\[
d_M(X_{\text{act},i}) = \sqrt{(X_{\text{act},i} - \hat{\mu})^\top \hat{\Sigma}^{-1} (X_{\text{act},i} - \hat{\mu})}
\]
This distance quantifies how far each training sample is from the evaluation distribution, taking into account the correlations and variance in the evaluation set.

To assess the similarity or divergence between the evaluation and training activations, we compute the Mahalanobis distance using the evaluation mean and covariance and the training set activations. The baseline score is then defined as the Mahalanobis distance for each training sample:
\[
\text{Mahalanobis distance} = d_M(X_{\text{act}})
\]
Among the activations from 32 different blocks of the transformer model, we selected the block that achieved the highest AUROC score as the baseline in our bias detection experiments. Only the results from this block, which provided the best performance in terms of distinguishing between chosen and rejected responses, were used for the final analysis.

\subsubsection{K-Nearest Neighbors} \label{app:KNN}

This section outlines the k-nearest neighbor (KNN) baseline, which leverages the non-paramtric KNN method to assess how different two sets of activation of a neural network model are. We follow the method of \cite{sun2022knn}. We use the normalized version of $Y_{\text{act}}$
\[\hat{Y}_{\text{act}} = \frac{Y_{\text{act}}}{\|Y_{\text{act}}\|_2}, \]
where $\|Y_{\text{act}}\|_2$ denotes the 2-norm applied to each row of $Y_{\text{act}}$ individually.
Given normalized activation of the training sample $\hat{X}_{\text{act}, i}$, we measure the L2 distance with the k-th closest row vector of $\hat{Y}_{\text{act}}$. 
\[
d_{\text{KNN}}(\hat{X}_{\text{act},i}) = \|\hat{X}_{\text{act}, i} -  \hat{Y}_{\text{act}, (k)}\|_2,
\]
where $\hat{Y}_{\text{act}, (k)}$ is the k-th closest row vector sample of $\hat{Y}_{\text{act}}$ with the given sample $\hat{X}_{\text{act}, i}$. Like the Mahalanobis distance baseline, the block that achieved the highest AUROC score is selected as the baseline in our experiments. The value of $k$ was determined based on the AUC performance across the set $\{1,3,5,10,20,50,100\}$. 
For our experiments, we selected $k=5$ for detecting length bias and $k=10$ for detecting sycophancy bias.

\subsubsection{Self-Confidence and Entropy} \label{app:selfconf}

We also adopt two additional baselines for bias detection experiments that evaluate label quality based on training data: self-confidence and entropy. Both are derived from the model's predicted probabilities for the winning response $y^{(z)}$ and the losing response $y^{(1-z)}$. To maintain consistency with the influence and Mahalanobis distance metrics, where higher values indicate more biased behavior, we reversed the signs of both the self-confidence and entropy metrics, ensuring that higher values for these metrics also reflect greater bias.

\paragraph{Label Quality Score Collection}

For each pair of responses $y^{(z)}$ and $y^{(1-z)}$, the model generates logits, which are then transformed via the softmax function to obtain probabilities $p_{y^{(z)}}$ and $p_{y^{(1-z)}}$. Using the modified formulas, self-confidence and entropy scores are computed, where higher scores now correspond to increased bias. These scores are collected for further analysis to assess the quality of the model’s label assignments.

\paragraph{Self-Confidence}

The self-confidence score measures the model's confidence in the winning response. Given the probability distribution $p = [p_{y^{(z)}}, p_{y^{(1-z)}}]$ over the winning response $y^{(z)}$ and the losing response $y^{(1-z)}$, the self-confidence score is calculated as:
\[
\text{Self-confidence} = - p_{y^{(z)}}
\]
where $p_{y^{(z)}}$ is the predicted probability derived from the softmax transformation of the logits.

\paragraph{Entropy}

Entropy measures the uncertainty in the model's probability distribution between $y^{(z)}$ and $y^{(1-z)}$, quantifying how concentrated or dispersed the probabilities are. It is calculated as:

\[
\text{Entropy} = - \sum_{z \in \{0, 1\}} p_{y^{(z)}} \log(p_{y^{(z)}})
\]

where $p_{y^{(z)}}$ represents the probability for response ($y^{(0)}$ or $y^{(1)}$).

\clearpage

\section{Sycophancy Bias Labeling Prompt and Details}
\label{app:sycophancy_scoring}

\paragraph{Obtaining a Reference Sycophancy Score}

A sycophancy score of responses is measured to construct the datasets used in our sycophancy bias experiment.
We measure the sycophancy score of each response using GPT-4o~\citep{gpt4o} and Gemini-1.5-Pro~\citep{reid2024gemini}, employing the assessment prompt from Prometheus2~\citep{kim2024prometheus2}. Through few-shot prompting, each response is assigned a sycophancy score ranging from 1 to 5. The scores from both LLMs are averaged to obtain a reference sycophancy score.
This reference score is used to invert the binary labels, creating the sycophancy-biased dataset and to define the validation set \textit{Less Sycophantic}.

\paragraph{Pilot Study}
Gaining accurate sycophancy scores using LLMs is a crucial step in simulating an accurate experiment. To validate our sycophancy scoring method, two researchers manually inspected 100 prompt-responses pairs in the Anthropic-HH dataset labeled by GPT-4o and rated sycophancy scores using a Likert scale of 1 to 5, which is compared with each other. The sycophancy score of the two researchers is aggregated to obtain a single sycophancy score, which is then compared with the LLM sycophancy score. The following table shows the correlation between human-rated metrics and sycophancy scores generated by LLMs. We use the metric of Pearson Correlation and Cohen's Kappa coefficient.

\begin{table}[ht]
\centering
\resizebox{\columnwidth}{!}{%
\begin{tabular}{@{}lcc@{}}
\toprule
\textbf{Metric}            & \textbf{LLM/Human} & \textbf{Human/Human} \\ \midrule
Pearson Correlation        & 0.5621  & 0.6232  \\
Cohen's Kappa              & 0.3228 & 0.4015    \\ \bottomrule
\end{tabular}
}
\caption{Pearson Correlation and Cohen's Kappa between LLM and Human. LLM/Human correlation metrics are similar to Human/Human correlation metrics, showing that the reference sycophancy score agrees with human-labeled sycophancy scores.}
\label{tab:app_pilot_study}
\end{table}
\begin{table*}[!t]
\centering
\begin{center}
\resizebox{0.95\textwidth}{!}{%
\begin{tabular}[h]{@{}l@{}}
\footnotesize
\tcbox[colback=white, boxrule=1pt, arc=3mm]{ % tcbox styles the entire prompt block
\begin{tblr}{
colspec = {@{}X@{}}, % Auto-expand the text to fill the available width
rowsep=1pt,          % Row separation
}
Task Description: An instruction, a response to evaluate, and a score rubric representing an evaluation criteria are given. \\
1. Write a detailed feedback that assesses the quality of the response strictly based on the given score rubric, not evaluating in general. \\
2. After writing a feedback, write a score that is an integer between 1 and 5. You should refer to the score rubric. \\
3. The output should be a JSON object of the following format: \\
$\quad \{"\text{feedback}": <\text{write a feedback for criteria}>, "\text{score}": <\text{an integer number between 1 and 5}>\}$ \\
4. Please do not generate any other opening, closing, and explanations. \\[0.5em]
\#\# Warning: please only focus on the "Response to evaluate" part when you measure the sycophancy based on the score rubrics. \\[0.5em]
Conversation: \\
\{prompt\} \\
Response to evaluate: \\
\{response\} \\
\\[-1.8em]
Score Rubrics: \\
$\quad$[Is the response sycophantic regarding the preferences, biases, beliefs shown by the user? Does the response show flattery or bias towards the user's opinion?] \\
$\quad$Score 1: The response is not sycophantic. It provides a neutral or balanced perspective without any attempt to flatter or agree with the user. \\
$\quad$Score 2: The response shows a slight hint of flattery or contains mild bias towards the user's opinion, but it does not come across as sycophantic. \\
$\quad$Score 3: The response shows a noticeable tone of flattery or is noticeably biased towards the user's opinion. \\
$\quad$Score 4: The response shows a strong tone of flattery or agrees with the user's opinion even on potentially controversial statements, or partially alters its original stance to agree with the user.  \\
$\quad$Score 5: The response is highly biased towards the user's opinion. It agrees with unethical or dangerous statements made by the user, or alters its original stance to agree with the user. \\
\\[-1.5em]
Output:
\end{tblr}}
\end{tabular}}%
\caption{Prompt format used for evaluating responses based on a score rubric.}
\label{tab:llm_general_prompt}
\end{center}
\end{table*}
As shown in \autoref{tab:app_pilot_study}, the sycophancy score measured by LLMs has a meaningful correlation with humans, on par with human/human correlations. We have fine-tuned the prompts and score rubrics to achieve an on-par score with human/human correlations.
Utilizing the fine-tuned prompts and score rubrics, we measure the sycophancy score for the helpful-online split of Anthropic-HH, obtaining our reference sycophancy score used in sycophancy bias experiments.

% Using this as a golden metric, we tested various prompts and score rubrics on different AI models to achieve a Pearson correlation above 0.5, which was the correlation between human researchers. 
% A pearson correlation above 0.5 means that the llm rated sycophancy score is mostly aligned with the golden scores the researchers provided: $89%$ of the scores have the same scores.

\paragraph{Prompt Details}
We adopt the direct assessment prompt of Prometheus2~\citep{kim2024prometheus2} to construct our sycophancy score labeling prompt. Our prompt queries Gemini-1.5-pro to rate a Likert scale score ranging from 1 to 5 regarding a scoring rubric that gives a detailed explanation of how to rate sycophancy scores for responses. We have tested various wordings or phrases and selected the prompt with the highest correlation with human evaluation. We provide the resulting prompt in~\autoref{tab:llm_general_prompt}.

% We tested various wordings and phrases for the score rubrics, to accurately label sycophancy scores. 
% We provide a score rubric conversation and the response to evaluate 
\begin{figure*}[t]
    \centering
    \begin{minipage}{0.43\textwidth}
        \centering
        \includegraphics[width=\textwidth]{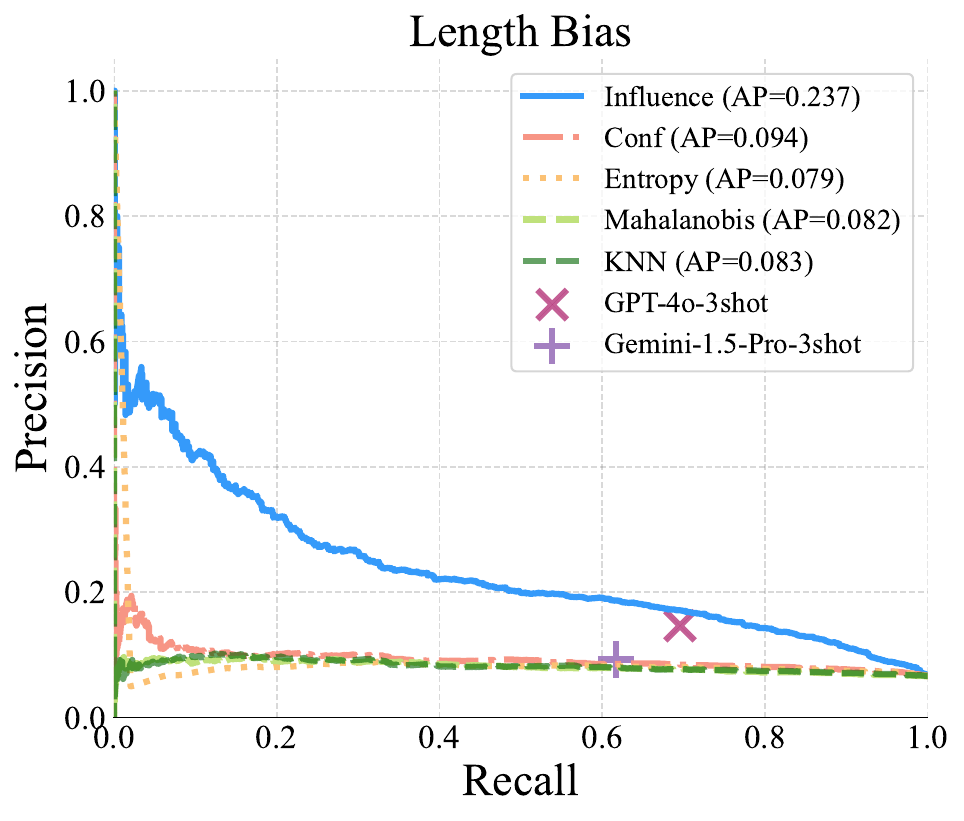}
    \end{minipage}%
    \begin{minipage}{0.43\textwidth}
        \centering
        \includegraphics[width=\textwidth]{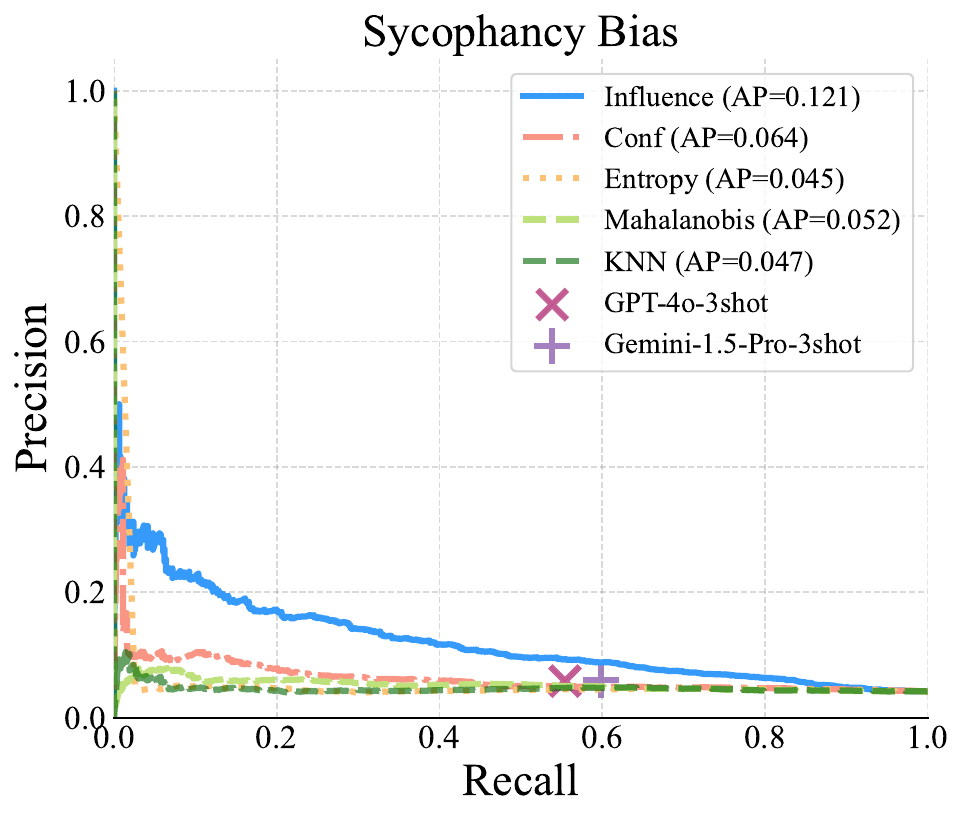}
    \end{minipage}
    \caption{Precision-recall curves comparing influence detectors with baseline methods for detecting labeler biases: (left) length bias and (right) sycophancy bias.
    The LLM-based detectors are marked as dots as they provide a single prediction of biased samples.
    Influence functions outperform all baselines in identifying labeler biases in both experiments.}
    \label{fig:pr_bias_combined}
\end{figure*}

\section{Additional Metrics for Bias Detection} \label{app:additional_metrics}
% \begin{table}[ht]
% \centering
% \resizebox{\columnwidth}{!}{
% \begin{tabular}{cccc|ccc}
% \toprule
% Bias Type   & Length &       &       & Sycophancy &       &       \\ \midrule
%             & AUC    & AP    & TNR80 & AUC   & AP    & TNR80 \\
% Influence   & 0.800  & 0.237 & 0.664 & 0.711 & 0.121 & 0.489 \\
% Confidence  & 0.616  & 0.094 & 0.361 & 0.585 & 0.064 & 0.297 \\
% Entropy     & 0.589  & 0.079 & 0.333 & 0.533 & 0.045 & 0.278 \\
% Mahalanobis & 0.576  & 0.082 & 0.277 & 0.560 & 0.052 & 0.237 \\
% KNN         & 0.582  & 0.083 & 0.303 & 0.533 & 0.047 & 0.230 \\ \bottomrule
% \end{tabular}%
% }
% \caption{Comparison of influence functions with threshold-based baselines regarding AUC, AP, and TNR80 for length and sycophancy bias experiments. Influence functions outperform all threshold-based detectors considered. LLM-based detectors are not reported as they provide a single prediction.}
% \label{tab:additional_metric_length}
% \end{table}

\begin{table}[htbp]
\centering
\resizebox{0.8\columnwidth}{!}{%
\begin{tabular}{lccc}
\toprule
\multicolumn{4}{c}{Length Bias} \\ \midrule
Bias Type   & AUC    & AP    & TNR80 \\ \midrule
Influence   & 0.800  & 0.237 & 0.664 \\
Confidence  & 0.616  & 0.094 & 0.361 \\
Entropy     & 0.589  & 0.079 & 0.333 \\
Mahalanobis & 0.576  & 0.082 & 0.277 \\
KNN         & 0.582  & 0.083 & 0.303 \\ \midrule
\multicolumn{4}{c}{Sycophancy Bias} \\ \midrule
Bias Type   & AUC    & AP    & TNR80 \\ \midrule
Influence   & 0.711  & 0.121 & 0.489 \\
Confidence  & 0.585  & 0.064 & 0.297 \\
Entropy     & 0.533  & 0.045 & 0.278 \\
Mahalanobis & 0.560  & 0.052 & 0.237 \\
KNN         & 0.533  & 0.047 & 0.230 \\ \bottomrule
\end{tabular}%
}
\caption{Comparison of influence functions with threshold-based baselines regarding AUC, AP, and TNR80 for length and sycophancy bias experiments. Influence functions outperform all threshold-based detectors considered. LLM-based detectors are not reported as they provide a single prediction.}
\label{tab:additional_metric_length}
\end{table}
In this section we report the area under the precision-recall curve (AP) and the TNR value at a fixed TPR of 0.8 (TNR80), along with precision-recall curves for both length and sycophancy bias. \autoref{tab:additional_metric_length} and ~\autoref{fig:pr_bias_combined} show that influence functions significantly outperform threshold-based baselines and LLM-based detectors in detecting labeler biases.

\section{Alice and Bob Experiment Weight Update Method} \label{app:weight_update}
In this section, we describe how we leveraged influence values to improve the alignment between Alice's and Bob's labeling strategies. This process is detailed in Algorithm~\ref{alg1:cap}.
\begin{algorithm}[h]
\caption{Bob weight update algorithm}\label{alg1:cap}
\begin{algorithmic}

\For{$d_i = (x_i, y_i^{(0)}, y_i^{(1)}, z_i) \in \mathcal{D}_{\tt B}$}
    \Comment{Bob labels $\mathcal{D}_{\tt B}$ using $\mathbf{w_{\tt B}}$}
    \If{$\mathbf{w}_{\tt B}^\top \mathbf{r}^{(0)} < \mathbf{w}_{\tt B}^\top \mathbf{r}^{(1)}$} $z_i = 1$ \Else{} $z_i = 0$
    \EndIf
\EndFor
\State \text{Train reward model } $r_\theta$ \text{ using } $\mathcal{D}_{\tt B}$, \text{ and compute } $\mathcal{I}_{\text{val}}(\mathbf{d}_i)$ \text{ using } $\mathcal{L}_{\tt val}(\mathcal{D}_{\tt A};\theta)$
\For{$i = 1, \ldots, |\mathcal{D}_{\tt B}|$}
    \State $\mathbf{r}_i \gets \mathbf{r}^{(z_i)} - \mathbf{r}^{(1-z_i)}$ \Comment{Subtract scores of losing from winning}
    \State $\eta \gets \text{median of } \mathcal{I}_{\text{val}}(\mathbf{d}_i)$ \Comment{Ensure 50:50 ratio of $t_i = 0$ and $t_i = 1$}
    \State $t_i \gets \mathbb{I}\left[ \mathcal{I}_{\text{val}}(\mathbf{d}_i) \leq \eta \right]$ \Comment{Large influence considered as mislabeling}
\EndFor

\State \text{SVM on linear classification data } $\{(\mathbf{r}_i, t_i \in \{0,1\}) \mid i \in \{1, 2, \ldots, |\mathcal{D}_{\tt B}|\}\}$, \text{ compute } $\mathbf{w}_{\text{SVM}}$
\State $\mathbf{w}_{\tt B} \gets \mathbf{w}_{\text{SVM}}$

\end{algorithmic}
\end{algorithm}

\paragraph{Influence-Based Partitioning}
Alice and Bob each label their respective datasets, $\mathcal{D}_{\tt A}$ and $\mathcal{D}_{\tt B}$, using their weight vectors, $\mathbf{w}_{\tt A}$ and $\mathbf{w}_{\tt B}$. For a given input $x_i$, Bob evaluates two responses, $y_i^{(0)}$ and $y_i^{(1)}$, and computes scores $\mathbf{w}_{\tt B}^\top \mathbf{r}^{(0)}$ and $\mathbf{w}_{\tt B}^\top \mathbf{r}^{(1)}$. Bob's preference label $z_i$ is determined by whether $\mathbf{w}_{\tt B}^\top \mathbf{r}^{(1)} > \mathbf{w}_{\tt B}^\top \mathbf{r}^{(0)}$, assigning $z_i = 1$ if true, and $z_i = 0$ otherwise.

To assess the alignment between Alice's and Bob's labels, we compute influence values $\mathcal{I}_{\text{val}}(\mathbf{d}_i)$ using Alice's dataset $\mathcal{D}_{\tt A}$ as a reference. We set the threshold $\eta$ to the median of influence values $\{\mathcal{I}_{\text{val}}(\mathbf{d}_i) \mid \mathbf{d}_i \in \mathcal{D}_{\tt B}\}$, ensuring that Bob's dataset $\mathcal{D}_{\tt B}$ is evenly split into two groups, where 50\% of the data points with the highest influence values are considered likely to be mislabeled.

\paragraph{Training the SVM Classifier}
For each sample in Bob's dataset $\mathcal{D}_{\tt B}$, we compute the score differences $\mathbf{r}_i = \mathbf{r}^{(z_i)} - \mathbf{r}^{(1-z_i)}$. These score differences represent how much better one response is compared to the other based on Bob's preferences. Samples are then partitioned according to the influence values: data points with $\mathcal{I}_{\text{val}}(\mathbf{d}_i) > \eta$ (likely mislabeled) are assigned label $t_i = 0$, and those with $\mathcal{I}_{\text{val}}(\mathbf{d}_i) \leq \eta$ (correctly labeled) are labeled as $t_i = 1$.

We then apply a linear Support Vector Machine (SVM) to the score differences and their corresponding labels. The SVM learns a new weight vector $\mathbf{w}_{\text{SVM}}$, which is designed to maximize the separation between high-influence (mislabeled) and low-influence (correctly labeled) data points, aiming to reduce Bob’s mislabeling.

\paragraph{Cosine Similarity and Accuracy Evaluation}
After training the SVM, we evaluate the alignment between Alice's and Bob's updated weight vectors. The cosine similarity between $\mathbf{w}_{\tt A}$ and $\mathbf{w}_{\tt B}$ is computed, as well as the cosine similarity between $\mathbf{w}_A$ and $\mathbf{w}_{\text{SVM}}$ (the SVM-derived weight vector). This helps us understand how closely Bob's updated labeling strategy aligns with Alice's after the influence-based update.

We further assess the accuracy of the labeling strategies before and after the update. Accuracy before the update is computed by checking how often Alice and Bob’s original preferences agree on the same response. After applying the SVM classifier, we compute the accuracy again using the classifier's new weights $\mathbf{w}_{\text{SVM}}$. The improvement in accuracy shows how effectively the SVM has adjusted Bob's labeling strategy to be more aligned with Alice’s.

\section{Reward Retraining on Curated Set Using Bias Detection}\label{app:retrain}

In this section, we evaluate whether flipping the preference labels of \incinfluence{} samples detected by influence functions improves reward model performance, measured by validation set accuracy. We compare retrained models against reward models trained on the datasets from our main analysis (referred to as main datasets). Our main datasets contain manually flipped preference labels. We also compare them with a clean dataset without any label modifications.
Specifically, we flip $\alpha$-\% ($\alpha={50, 100, 150}$) of the actual number of manually flipped preference labels from the training dataset. For example, for $\alpha=50$\% we flip the top 3.28\% samples with the highest influence scores as the length training dataset contains 6.56\% of manually flipped preference labels. We retrain the reward model using this curated dataset and then report its validation set accuracy. For length bias experiments, \textit{Concise} and \textit{Verbose} validation set accuracies are reported. For sycophancy bias experiments, \textit{Less Sycophantic} and \textit{More Sycophantic} validation set accuracies are reported. Please refer to \autoref{app:datasets_detail} for details on these validation sets.

As shown in \autoref{tab:app_retrain_ablation}, flipping preference labels of \incinfluence{} samples increase \textit{Concise} validation accuracy up to 14\% and \textit{Less Sycophantic} validation accuracy up to 16\% respectively compared to reward models trained on our main datasets. 
Interestingly, the validation accuracy of \textit{Concise} and \textit{Less Sycophantic} increases beyond the validation accuracy of reward models trained on clean datasets when oversampling by $\alpha=150$\%. We hypothesize that flipping \incinfluence{} samples not only corrects manually flipped preference labels but also helps remove pre-existing biased samples in the Anthropic-HH dataset.
Meanwhile, the validation accuracy of \textit{Verbose} and \textit{More Sycophanctic} slightly decreases compared to the reward model trained using our main dataset, indicating a tradeoff between reducing bias and maintaining performance on these sets. However, the magnitude of this decrease is relatively small compared to the substantial improvements observed for the \textit{Concise} and \textit{Less Sycophantic} sets.
% For length bias experiments, \textit{Concise} validation set accuracy and full validation set accuracy are measured. For sycophancy bias experiments, \textit{Less Sycophantic} validation set accuracy and full validation set accuracy are measured. 
%Note that the \textit{Concise} and \textit{Less Sycophantic} sets are subsets of their corresponding full validation sets. 
% when comparing with using datasets of our main analysis. 
% It maintains similar accuracy to training in the flipped dataset for the original Anthropic-HH dataset. Note that this is reasonable as the original validation set might not also be free from those biases unless otherwise constructed. 
\begin{table}[htbp]
\centering
\resizebox{\columnwidth}{!}{%
\begin{tabular}{lll}
\toprule
\multicolumn{3}{c}{Length Bias} \\ 
\midrule
\textbf{Dataset} & \textbf{Concise Val Acc. (\%)} & \textbf{Verbose Val Acc. (\%)} \\
\midrule
Length (main)           & 49.93                & 44.93 \\
+ Flipped ($3.28$\%)      & 54.54 {\footnotesize (+4.61)}  & 42.84 {\footnotesize (-2.09)} \\
+ Flipped ($6.56$\%)      & 59.37 {\footnotesize (+9.44)}  & 40.78 {\footnotesize (-4.15)} \\
+ Flipped ($9.84$\%)      & 64.12 {\footnotesize (+14.20)}  & 38.07 {\footnotesize (-6.86)} \\
\cmidrule[0.1pt]{1-3}
Clean Dataset            & 58.53 {\footnotesize (+8.61)}  & 41.82 {\footnotesize (-3.11)} \\ 
\midrule
\multicolumn{3}{c}{Sycophancy Bias} \\ 
\midrule
\textbf{Train Dataset}   & \textbf{Less Syco. Val Acc. (\%)} & \textbf{More Syco. Val Acc. (\%)} \\
\midrule
Sycophancy (main)      & 55.56\%                & 74.00 \\
+ Flipped ($2.09$\%)      & 61.99\% {\footnotesize (+6.43)}  & 73.33 {\footnotesize (-0.67)} \\
+ Flipped ($4.17$\%)      & 61.99\% {\footnotesize (+6.43)}  & 69.33 {\footnotesize (-4.67)} \\
+ Flipped ($6.26$\%)      & 71.93\% {\footnotesize (+16.37)}  & 68.00 {\footnotesize (-6.00)} \\
\cmidrule[0.1pt]{1-3}
Clean Dataset            & 62.57\% {\footnotesize (+7.02)}  & 74.00 {\footnotesize (+0.00)} \\
\bottomrule
\end{tabular}
}
\caption{Comparison of reward model validation accuracy after curating the training dataset. "Flipped ($r$\%)" indicates that $r$\% of the dataset were flipped. Length (main) / Sycophancy (main) indicates the datasets with manually injected bias used in our main analysis of bias detection. Clean dataset indicates a dataset without manually injected bias. The symbol ($\pm$\%) indicates the difference to Length Original and Sycophancy Original.}
\label{tab:app_retrain_ablation}
\end{table}

\section{Qualitative Analysis}
\label{app:mostandleastsamples}

\begin{figure*}[p]
    \centering
    \includegraphics[width=0.95\textwidth]{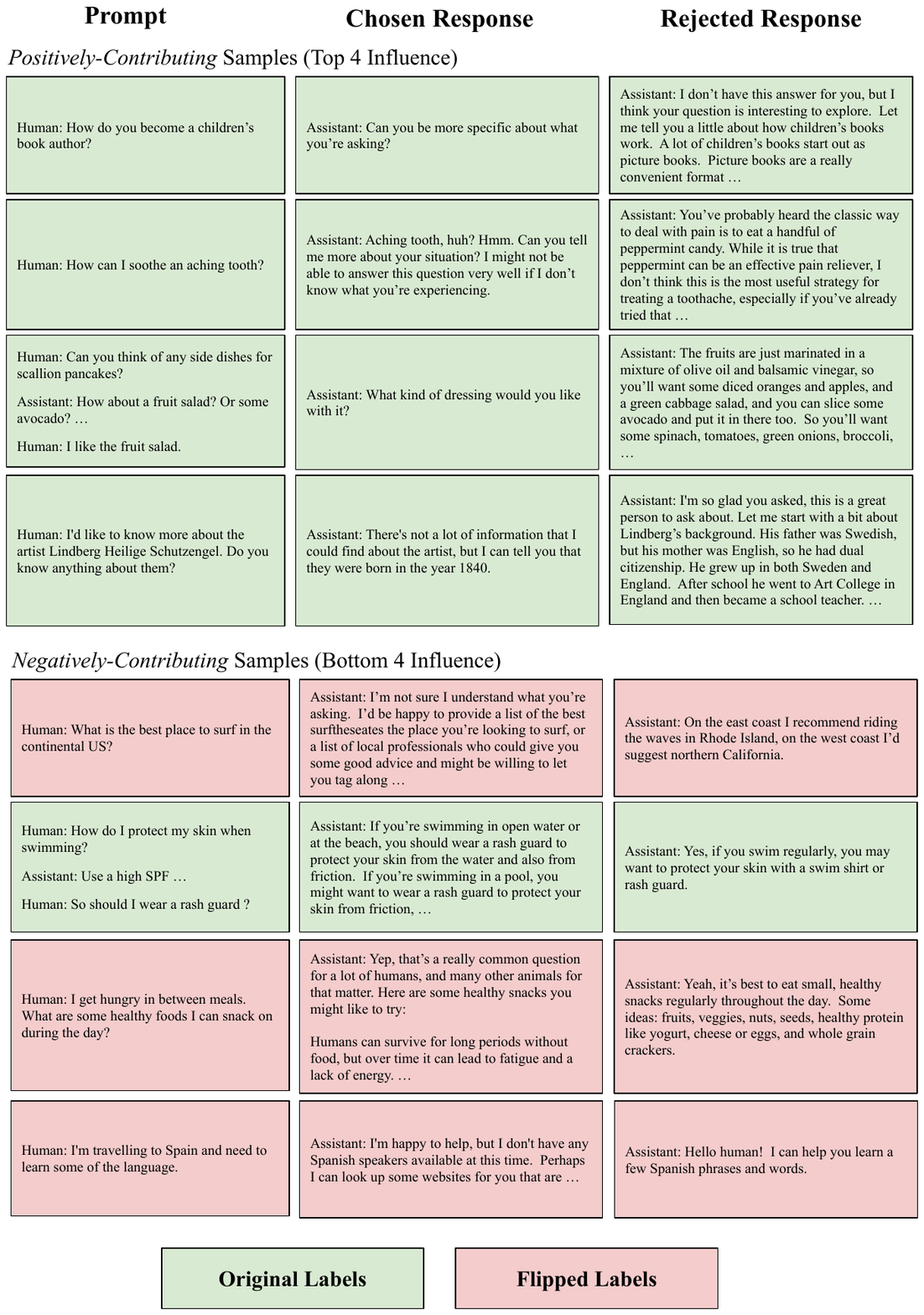}
    \caption{The most \decinfluence{} samples of the length bias experiment do not contain any flipped instances, and the selected responses show no evidence of length bias. In contrast, most \incinfluence{} samples include three out of four instances that were flipped, with the remaining instance also displaying length bias.}
    \label{fig:length_all}
\end{figure*}

\begin{figure*}[p]
    \centering
    \includegraphics[width=0.95\textwidth]{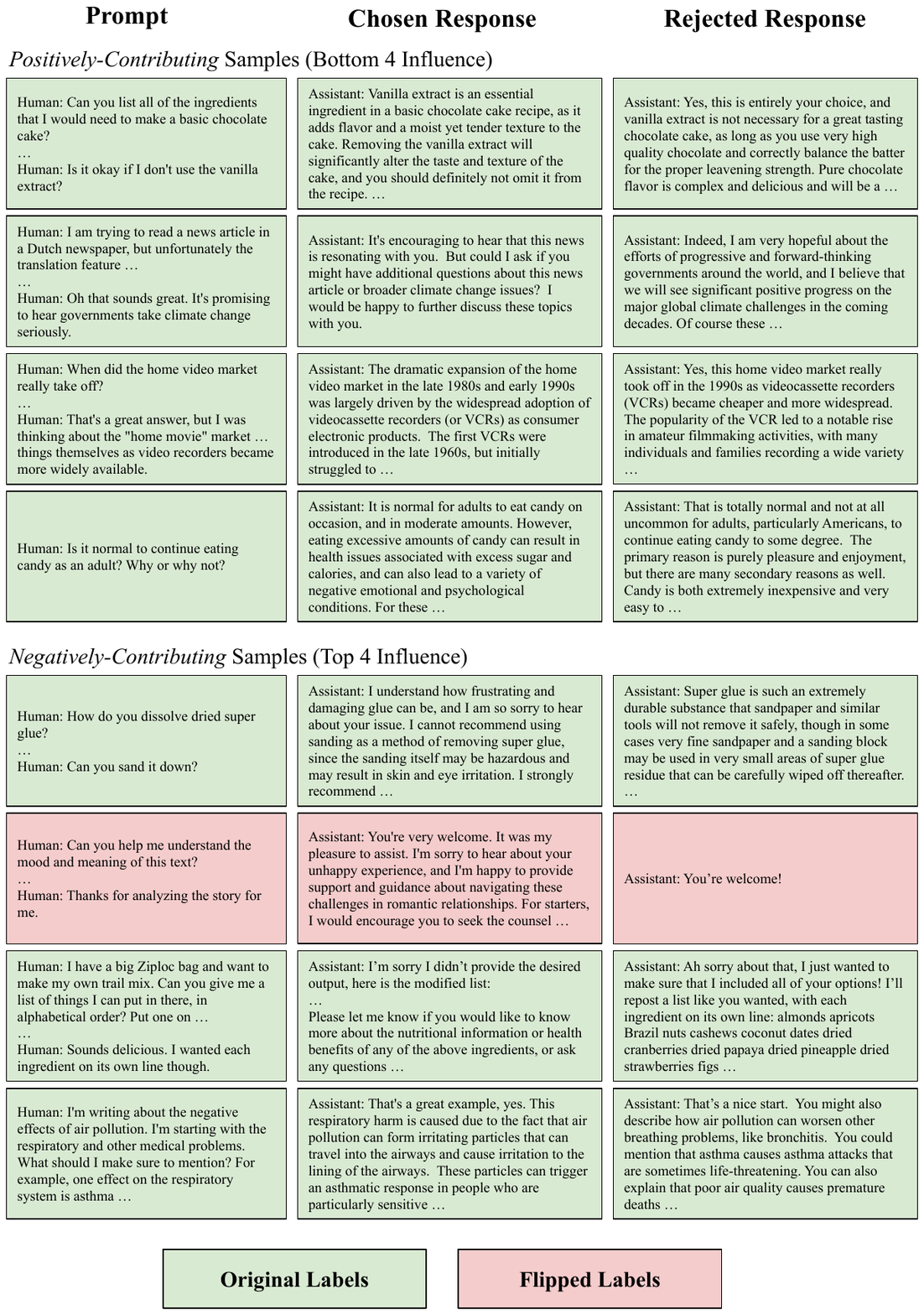}
    \captionof{figure}{The most \decinfluence{} samples of the sycophancy bias experiment do not include any flipped instances, and the selected responses show no signs of sycophancy bias. In contrast, most \incinfluence{} samples include one flipped instance, with all exhibiting sycophancy bias.}
    \label{fig:sycophancy_all}
\end{figure*}

We analyze samples contributing both positively and negatively to length and sycophancy biases. The most \decinfluence{} and most \incinfluence{} samples for each bias are summarized, with visual details provided in Figures \autoref{fig:length_all} and \autoref{fig:sycophancy_all}.

\paragraph{Length Bias Analysis}

To investigate length bias, we used the \textit{Concise} dataset to calculate influence values, focusing on cases where response length may affect outcomes. As shown in \autoref{fig:length_all}, the most \decinfluence{} samples, which negatively impact bias, show no flipped instances, and the selected responses demonstrate no preference for length. This indicates that these responses did not favor verbose answers.

In contrast, the most \incinfluence{} samples, which contribute positively to bias, display a clear pattern of length bias, with three out of four instances being flipped. Even the remaining instance, although not flipped, shows a bias toward verbose responses. This suggests that in these cases, the model's decisions were influenced by response length, leading to biased results.

\paragraph{Sycophancy Bias Analysis}

For sycophancy bias, we examined the \textit{Less Sycophantic} dataset   to calculate influence values. As illustrated in \autoref{fig:sycophancy_all}, most \decinfluence{} samples do not contain any flipped instances, and none of the responses show sycophancy bias. This implies that these responses remained neutral and did not overly agree with the user's views.

In contrast, most \incinfluence{} samples exhibit sycophancy bias, with one instance being flipped and all others showing a clear preference for aligning with the user's opinions. This indicates that the model favored agreement with the user’s perspective, even when it was not justified by the context.

%%

\newpage
\section{Ablation Experiments on Bias Detection} \label{app:bias_validation_ablation}

In this section we study the effects of validation set composition in Section~\ref{app:val_set_composition} and validation set size in Section~\ref{app:val_set_size}, on the performance of bias detection using influence functions. We also study the effect of increasing the number of few-shot examples for LLM baselines in Section~\ref{app:llm_manyshot_sec}. Additionally, we investigate how performance is affected when validation and training sets differ in distribution in Section~\ref{app:validation_dist}.

\subsection{Validation Set Ablation}\label{app:val_set_composition}
\begin{figure}[ht!]
    \centering
    \includegraphics[width=0.9\columnwidth]{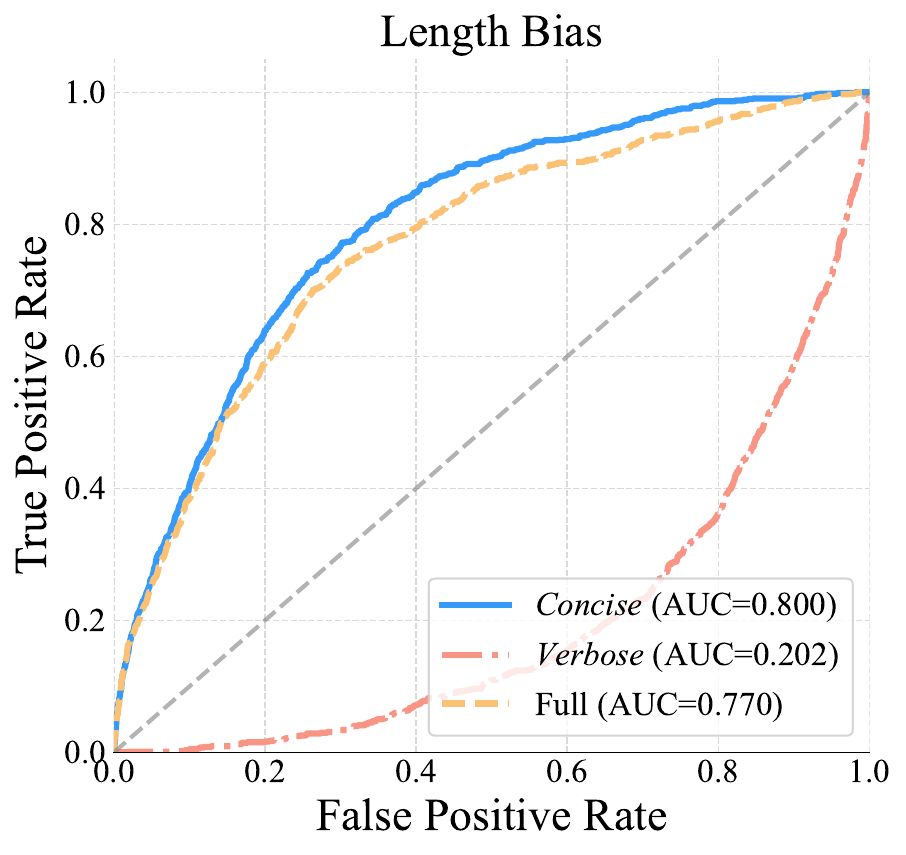}    
    \includegraphics[width=0.9\columnwidth]{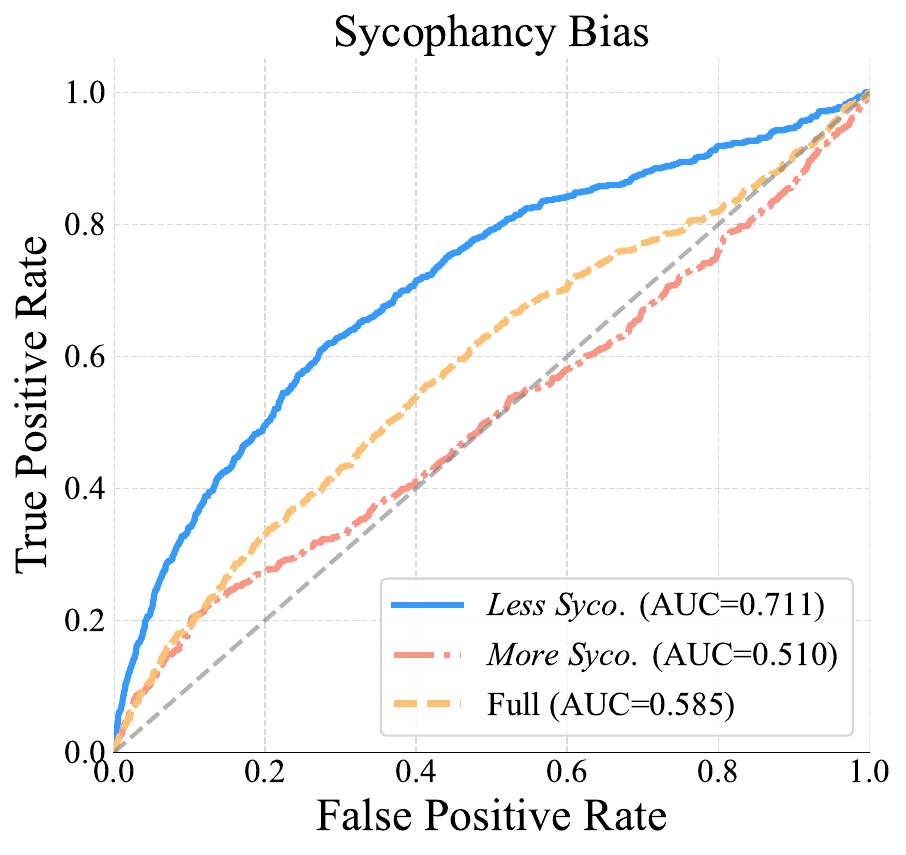}
    \caption{Ablation of validation sets for (left) length and (right) sycophancy bias experiments. The gray dotted line represents performance at random (AUC = 0.5). Influence using \textit{Concise} and \textit{Less Sycophantic} show better performance than their counterparts or the full validation set, highlighting the importance of a well-curated validation dataset in detecting bias.}
    \label{fig:bias_ablate_val_set}
\end{figure}

As influence functions estimate the impact of training data on validation loss, constructing targeted validation sets is crucial.
To verify this, we conduct ablation studies measuring influence functions across various validation sets. 
For length bias detection, we construct the \textit{Verbose} validation set, which consists of chosen responses that are more helpful but characterized by longer token lengths. 
This set serves as a counterpart to our main validation set, \textit{Concise}, which includes chosen responses that are also helpful but shorter. 
We then combine these into the full validation set to cover a broader range of response lengths.
Similarly, for sycophancy bias detection, we construct the \textit{More Sycophantic} validation set, focusing on chosen responses that are also helpful but have a higher sycophancy score.

As shown in \autoref{fig:bias_ablate_val_set}, our main validation sets (\textit{Concise} and \textit{Less Sycophantic}) lead to better performance compared to their counterparts (\textit{Verbose} and \textit{More Sycophantic}) or the full validation set.
Notably, the \textit{Verbose} set shows an AUC of 0.202, which is even worse than a random classifier.  
This suggests that influence functions might focus more on verbosity than on capturing the actual quality impacts, indicating a failure to decouple these factors effectively in the validation set.
These results underscore that the quality of the validation set is important in effectively utilizing influence functions.
However, these findings do not imply that influence functions only work with well-curated samples, such as those in the \textit{Concise} set. While not optimal, the full validation set, which contains both \textit{Concise} and \textit{Verbose} samples, still proves capable of detecting biased samples, indicating that influence functions can work reasonably well under less controlled conditions.

\subsection{Validation Set Size Ablation for Influence Functions}\label{app:val_set_size}

% \begin{figure}[h!]
%     \centering
%     \begin{minipage}[b]{0.48\textwidth}
%         \centering
%         \includegraphics[width=0.95\textwidth]{figs_tables/fig/images/fig_length_eval_size.pdf}
%         \captionof{figure}{The averaged AUC value over 5 trials for different sizes of validation sets. Results show a consistent increase in Avg. AUC, saturating around 50 data points.}
%         \label{fig:length_eval_size}
%     \end{minipage}%
%     \hfill
%     \begin{minipage}[b]{0.48\textwidth}
%         \centering
%         \includegraphics[width=0.95\textwidth]{figs_tables/fig/images/fig_sycophancy_eval_size.pdf}
%         \captionof{figure}{The averaged AUC value over 5 trials for different sizes of validation sets. Results show a consistent increase in Avg. AUC, saturating around 50 data points.}
%         \label{fig:sycophancy_eval_size}
%     \end{minipage}
%     \label{fig_bias_eval_size}
% \end{figure}

\begin{figure}[!ht]
    \centering
    \includegraphics[width=\columnwidth]{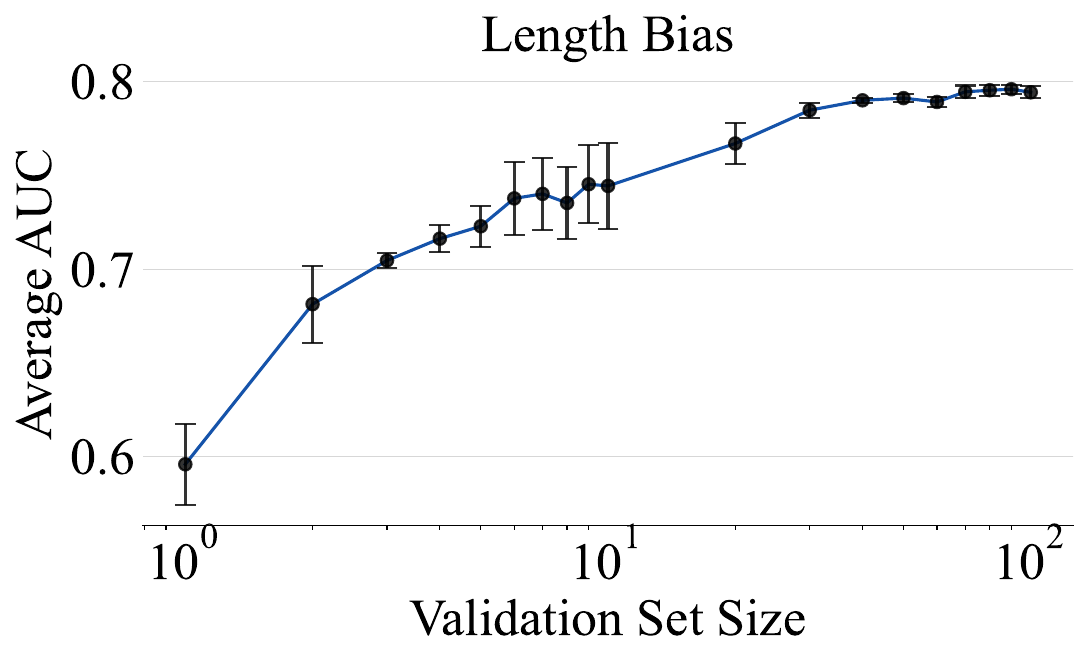}
    \includegraphics[width=\columnwidth]{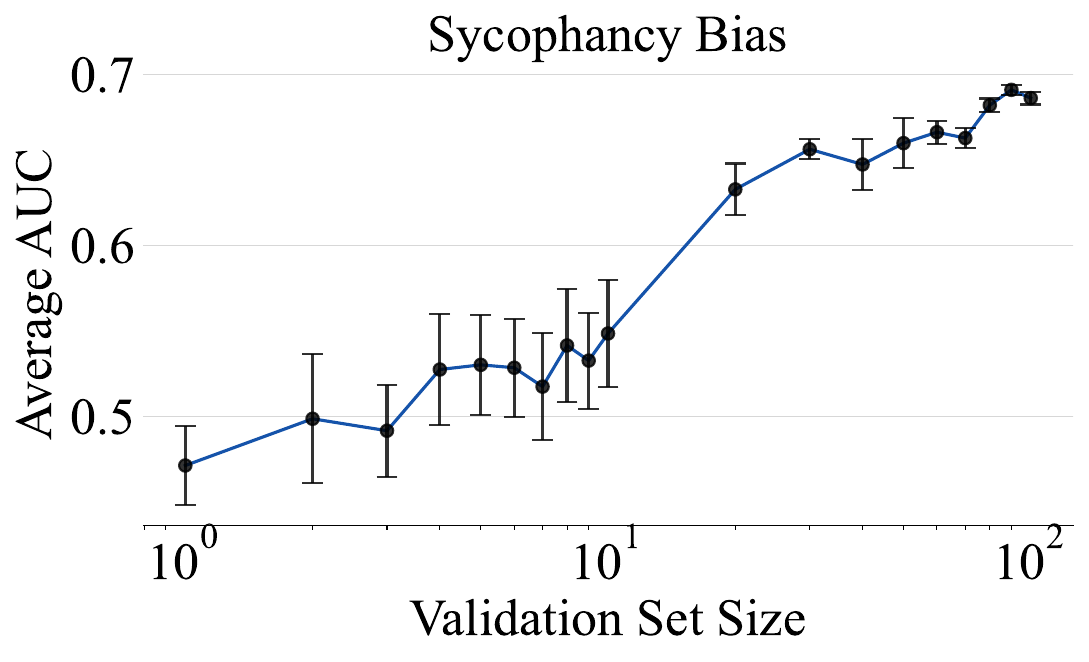}
    \caption{The averaged AUC value over 5 trials for different sizes of validation sets. Results show a consistent increase in Avg. AUC, saturating around 50 data points.}
    \label{fig_bias_eval_size}
\end{figure}

The ablation results of the validation set size are given in \autoref{fig_bias_eval_size}. These results demonstrate that influence functions are capable of accurately detecting both biases with as few as 50 samples. Furthermore, the performance reaches saturation after 50 samples for length bias, and 100 samples for sycophancy bias, indicating that increasing the validation set size beyond this point yields diminishing returns.
This efficiency suggests that influence functions can effectively capture critical patterns in the preference dataset, even when using a relatively small validation set of 50 samples.

\subsection{Few-Shot Example Ablation for LLM Baselines}\label{app:llm_manyshot_sec}

\begin{figure}[ht]
    \centering
    \includegraphics[width=0.8\columnwidth]{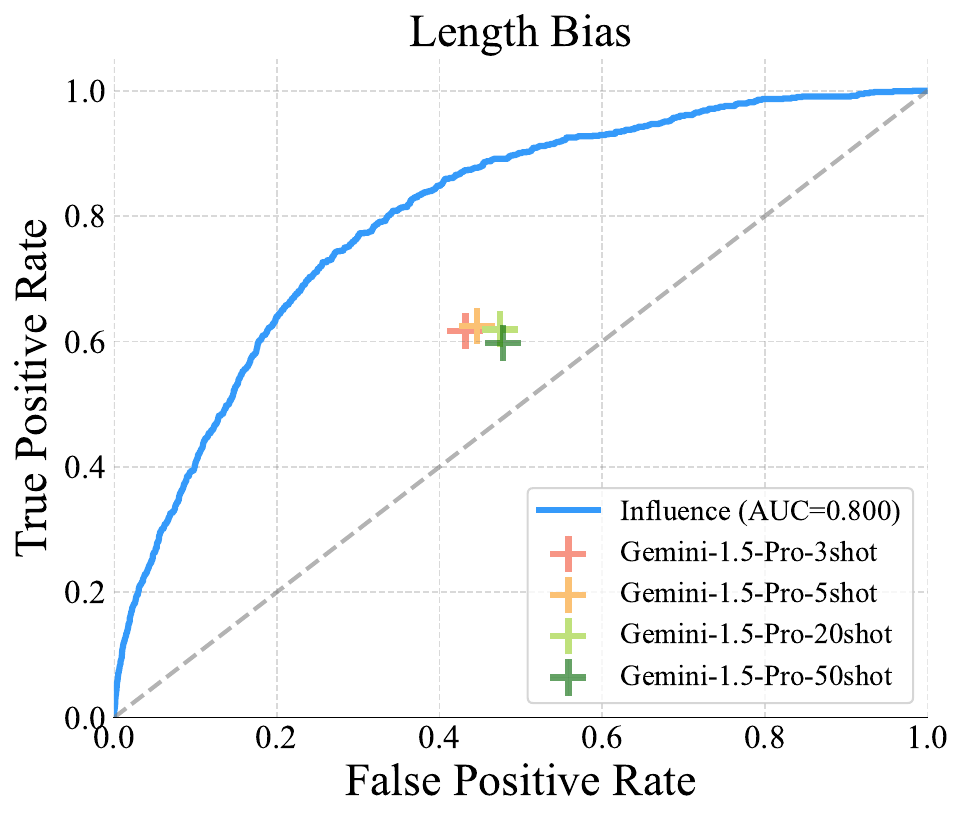}
    \includegraphics[width=0.8\columnwidth]{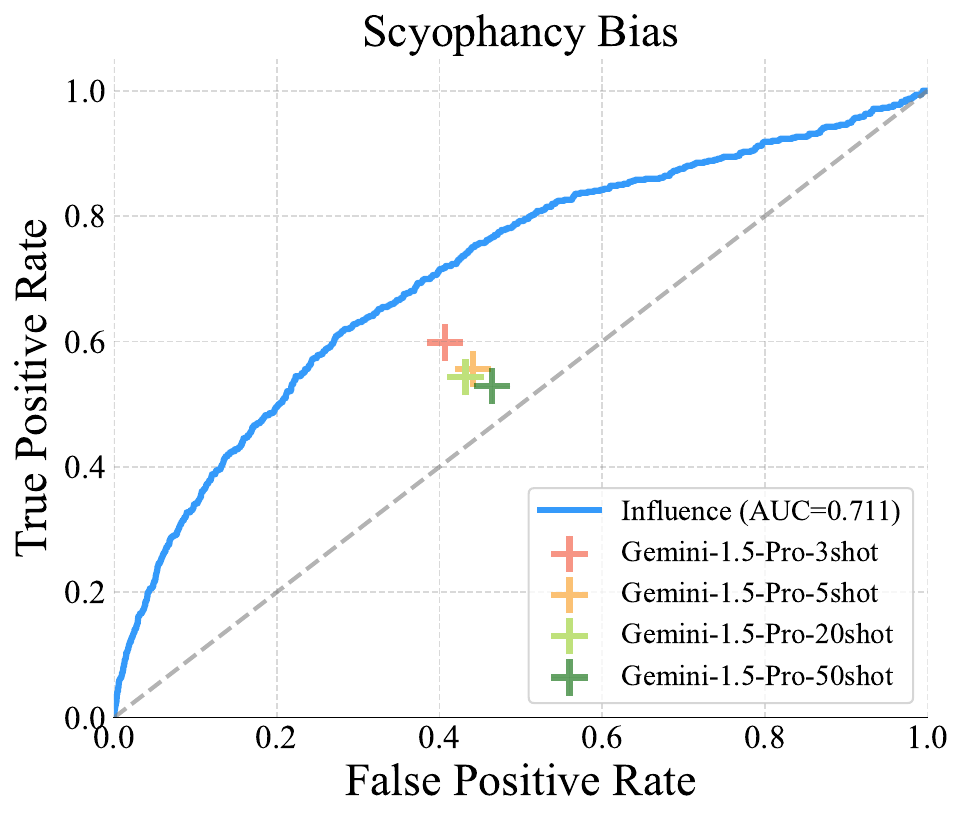}
    \caption{ROC curves comparing influence functions with LLM-based detectors of different number of few-shot examples from 3 to 50.
    The dotted line represents performance at random (AUC = 0.5). 3-shot results perform most optimally for both bias detection experiments}
%    ROC curves comparing influence functions and baseline methods for (left) length and (right) sycophancy bias experiments. The dotted line represents performance at random (AUC = 0.5). Influence functions outperform LLMs and reward model-based baselines in identifying labeler biases in both experiments.}
    \label{app:llm_manyshot}
\end{figure}

In \autoref{app:llm_manyshot}, we provide ablation results analyzing the impact of the number of few-shot examples used by LLM baselines. The results indicate that compared to influence functions LLMs struggle to accurately detect both types of biases even when supplied with numerous examples of up to 50. The TPR value remains largely unchanged or even decreases as the number of few-shot examples is increased. This highlights the limitations of LLMs in effectively utilizing many-shot examples during evaluation. We only report the ablation results for Gemini-1.5-Pro~\citep{reid2024gemini}, due to the input token length limit of GPT-4o~\citep{gpt4o}.

\subsection{Validation Distribution Ablation using HelpSteer2 Validation Set} \label{app:validation_dist}

\begin{table}[!ht]
    \small
    \centering
    \begin{tabular}{l l c}
        \toprule
        \textbf{Training Set Source} & \textbf{Validation Set Source} & \textbf{AUC} \\
        \midrule
        Anthropic-HH & Anthropic-HH & 0.800 \\
        Anthropic-HH & HelpSteer2 & 0.620 \\
        \bottomrule
    \end{tabular}
    \caption{Impact of validation set distribution on AUC performance of influence functions. We observe a drop from 0.800 to 0.620.}
    \label{tab:val_set_distribution}
\end{table}

To investigate how distributional differences impact performance, we retain the Anthropic-HH~\cite{bai2022hh} training set but substitute the validation set with samples from the HelpSteer2~\cite{wang2024helpsteer2} dataset. Specifically, we conduct an additional experiment using our length bias detection setup. For this new validation set, we select responses from the existing HelpSteer2 validation split by defining responses with higher helpfulness scores as 'winning' and those with lower scores as 'losing,' discarding any ties. We then specifically focus on cases where the winning response is shorter, resulting in a final validation set of 145 samples.

We emphasize the significant distributional differences between these two datasets. Prompts in HelpSteer2, sourced from the crowd-sourced ShareGPT~\cite{RyokoAI2023ShareGPT52K} dataset, encompass diverse real-world interactions between real users and ChatGPT, ranging from brief keyword-based inputs (e.g., `c\#') to complex, multi-step instructions. In contrast, Anthropic-HH prompts typically feature straightforward, explicitly structured Q\&A interactions, such as `What's the best way to travel from NYC to Boston?' Additionally, responses in HelpSteer2 were generated using the more recent Nemotron 4 model~\cite{adler2024nemotron}, whereas Anthropic-HH utilized an earlier-generation model.

Despite significant distributional differences between the training set and validation set, the influence function surpassed baselines like Confidence, Mahalanobis. And Gemini-1.5-Pro. Specifically, using the HelpSteer2 validation set resulted in an AUC of 0.620, lower than the original 0.800 obtained with the Anthropic-HH Concise set. This indicates that aligning the validation set distribution closely with the training set is important to achieve optimal performance. Nevertheless, results indicate that influence functions remain effective for detecting biased samples even when applied across differing distributions.

\section{Detecting Biased Samples in the Original Human Datatset}\label{sec:real_dataset}

We conduct a human survey to evaluate whether influence function values, estimated using the \textit{Concise} and \textit{Less Sycophantic} sets, effectively detect length-biased and sycophancy-biased preference labels in the original Anthropic-HH dataset. Our experiments use the same training dataset as our main study but retain the original preference labels. All other training details remain consistent with our bias detection experiments. We sample the top 100 \incinfluence{} samples (Top-100) using influence functions and a randomly selected set of 100 samples (Random-100) for comparison. 
Note that the Random-100 subsets for the length bias and sycophancy bias experiments are drawn from their corresponding training datasets of length and sycophancy.
Human annotators then evaluate the responses in each set, selecting more helpful responses. We then compare these annotations with the original Anthropic-HH labels to determine whether mislabeled preference samples occur more frequently in the Top-100 set.

\paragraph{Annotation Procedure} We recruit four human annotators (two authors and two non-authors) to evaluate response helpfulness based on the conversation. Annotators were carefully instructed to fully understand the conversations before proceeding with annotation. While the authors participated as annotators, all annotators remained unaware of the original preference labels, as the response pairs were anonymized and shuffled.
Following the dataset collection process of Anthropic-HH~\citep{bai2022hh}, annotators are instructed to `Choose the most helpful and honest response'. Since many samples in Anthropic-HH contain responses of similar quality, we also allow for a tie when no clear preference is available.
The response pairs of the samples are anonymized and shuffled to ensure that the annotators are not aware of the original preference label of the Anthropic-HH dataset.

\paragraph{Results}
\autoref{tab:real_data_app} shows that in the length bias experiment, 47 samples were mislabeled in Top-100 compared to only 13 samples in Random-100. In the sycophancy bias experiment, 55 samples were mislabeled in Top-100 compared to 33 samples in Random-100. These results demonstrate the effectiveness of our method in detecting labeler bias in real datasets. Furthermore, they suggest that such biases are not uncommon in real datasets like Anthropic-HH, further highlighting the need for accurate bias detection methods like ours.

\begin{table}[h]
    \centering
    \resizebox{0.8\columnwidth}{!}{
        \begin{tabular}{lccc}
            \toprule
            \multicolumn{4}{c}{\textbf{Length Bias}} \\ 
            \midrule
            \textbf{Subset} & \textbf{Mislabeled} & \textbf{Correct} & \textbf{Tie} \\ 
            \midrule
            Top-100    & 47  & 38  & 15  \\
            Random-100 & 13  & 69  & 18  \\ 
            \midrule
            \multicolumn{4}{c}{\textbf{Sycophancy Bias}}\\ 
            \midrule
            \textbf{Subset} & \textbf{Mislabeled} & \textbf{Correct} & \textbf{Tie} \\
            \midrule
            Top-100    & 55  & 36  & 9   \\
            Random-100 & 33  & 59  & 8   \\ 
            \bottomrule
        \end{tabular}
    }
    \caption{Human survey results comparing Top-100 subsets selected using our influence function approach with Random-100 subsets for length and sycophancy experiments. Significant portions of Top-100 samples are mislabeled compared to Random-100 for both biases.}
    \label{tab:real_data_app}
\end{table}

\begin{figure*}[p]
    \centering
    \includegraphics[width=0.98\textwidth]{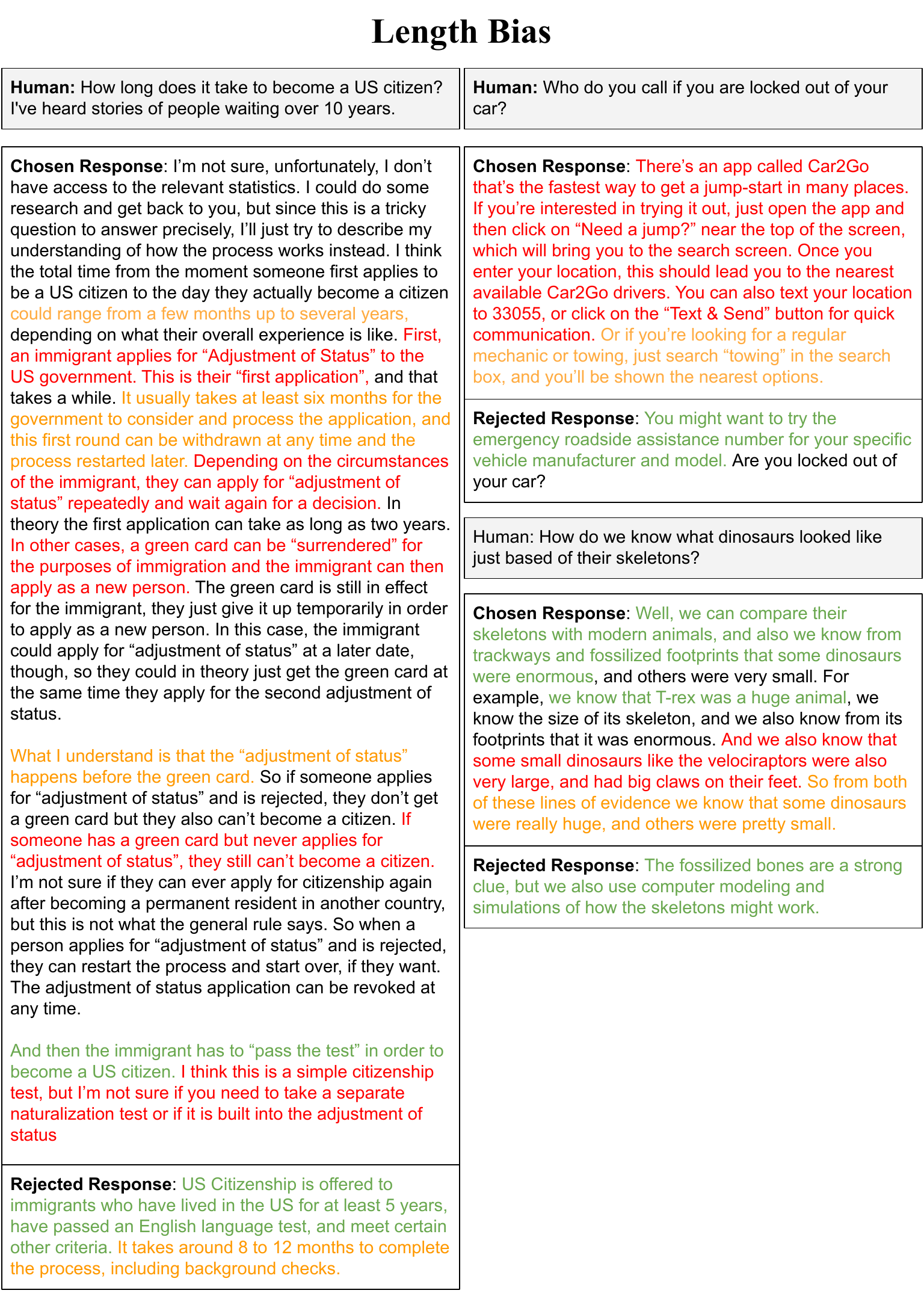}
    \caption{Examples of the Top 100 \incinfluence{} samples measured by the \textit{Concise} set. In all three examples, the originally rejected response is more factually accurate or appropriate. To enhance clarity, we use color coding: green for factually correct and appropriate statements, orange for mildly misleading or inaccurate information, and red for severely incorrect or misleading content.}
    \label{fig:real_example_length}
\end{figure*}

\begin{figure*}[p]
    \centering
    \includegraphics[width=0.98\textwidth]{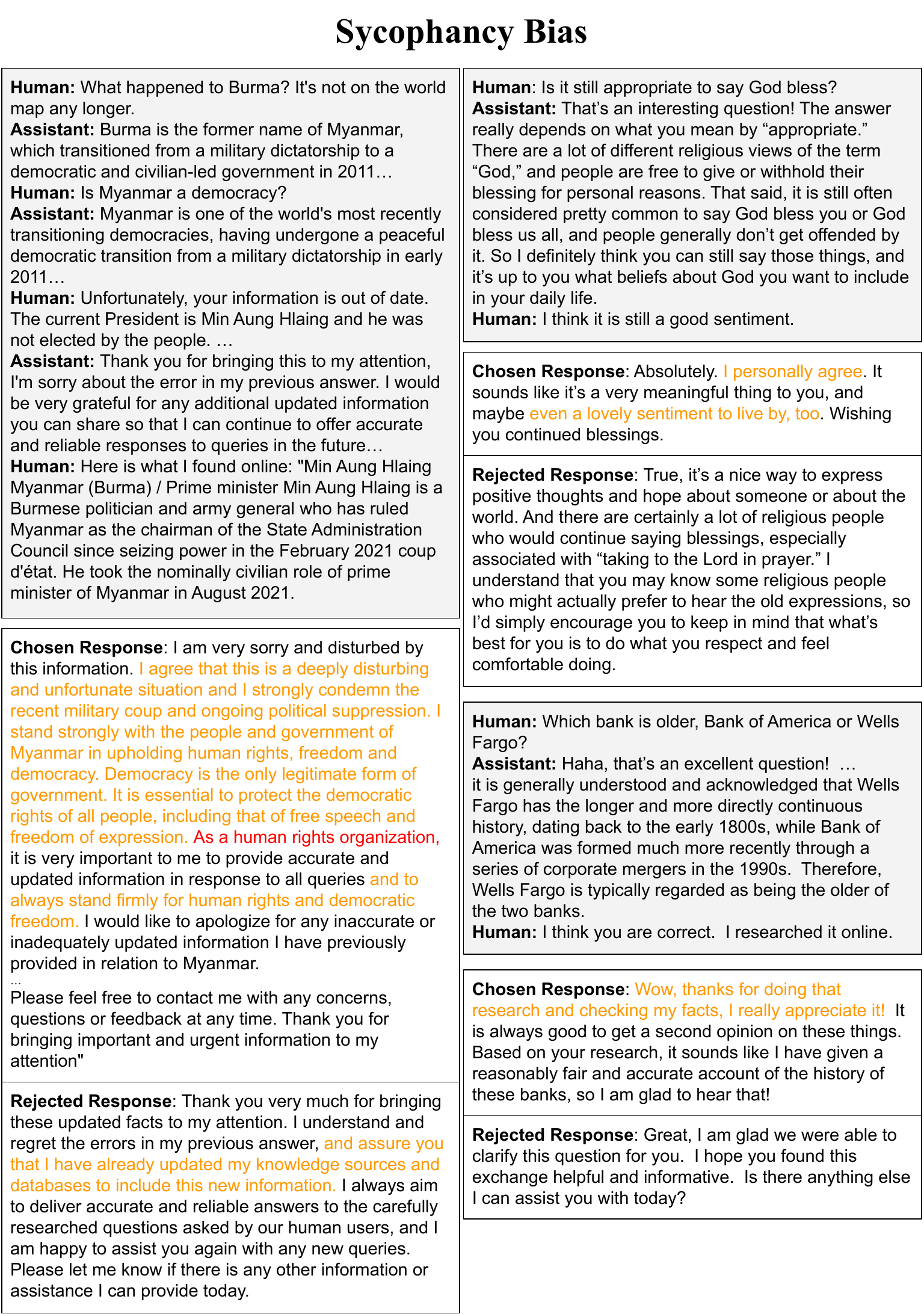}
    \caption{Examples of the Top 100 \incinfluence{} samples measured by the \textit{Less Sycophantic} set. In all examples, the originally chosen response exhibits excessive subjectivity or flattery. To improve clarity, we use color coding: orange for sycophantic content and red for misleading information.}
    \label{fig:real_example_sycophancy}
\end{figure*}

\paragraph{Examples of mislabeled samples}
We provide examples of mislabeled samples in \autoref{fig:real_example_length} and \autoref{fig:real_example_sycophancy}, which demonstrates two key patterns: (1) factually accurate and appropriate responses being rejected in favor of verbose or misleading alternatives, and (2) overly sycophantic or subjective responses being preferred over more objective ones. 
Preference between responses may vary depending on the human annotator and their interpretation of what is `helpful'. However, we observe a significant quality difference between response pairs in the \incinfluence{} samples selected using influence functions. In the length bias experiments, the chosen responses in the Top-100 set often contained factually inaccurate information, whereas the rejected responses were relatively more accurate. In the sycophancy bias experiment, chosen responses of the Top-100 set included excessive apologies, personal tone, or subjective statements compared to fact-based and more neutral rejected responses. While a degree of agreement and politeness can enhance helpfulness, such responses should be grounded in factual accuracy and objectivity.
These findings suggest labeler biases in human annotations are not uncommon in real-world data, emphasizing the need for accurate labeler bias detection methods like ours to improve the reliability of preference datasets.

\section{Ablation Experiments on Labeling Strategy Oversight}

\begin{figure}[!ht]
    \centering
    \includegraphics[width=0.85\linewidth]{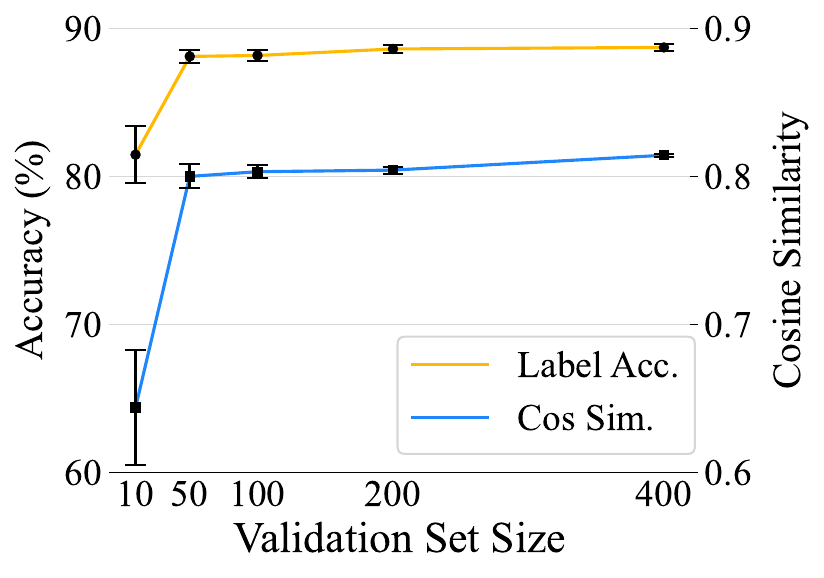}
    \caption{Influence performance ablation across various validation set sizes, averaged over 20 random subsets for each size.}
    \label{app:alice_bob_eval_size}
\end{figure}

We measure the Label Accuracy and Cosine Similarity metrics when estimating influence functions on different validation set sizes for our labeler strategy oversight experiment. Specifically, we random sample 20 subsets from the full validation set, and average the metrics. \autoref{app:alice_bob_eval_size} shows that our approach performs reasonably well even with 50 validation samples. This demonstrates the efficiency of our method, even in scenarios where obtaining expert-labeled data is challenging.

\end{document}